\documentclass[letterpaper]{article} 
\usepackage{aaai25}  
\usepackage{times}  
\usepackage{helvet}  
\usepackage{courier}  
\usepackage[hyphens]{url}  
\usepackage{graphicx} 
\usepackage{array}
\usepackage{booktabs}
\usepackage{multirow}
\usepackage{subcaption}
\usepackage{tcolorbox}
\usepackage{tabularx}
\usepackage{amsmath}
\usepackage{colortbl}
\usepackage[utf8]{inputenc}
\usepackage[greek,english]{babel}

\definecolor{lgray}{rgb}{0.9,0.9,0.9}

\def\UrlFont{\rm}  
\usepackage{natbib}  
\usepackage{caption} 
\usepackage{algorithm}
\usepackage{algorithmic}

\usepackage{newfloat}
\usepackage{listings}
\DeclareCaptionStyle{ruled}{labelfont=normalfont,labelsep=colon,strut=off} 
\floatstyle{ruled}
\newfloat{listing}{tb}{lst}{}
\floatname{listing}{Listing}
\title{What can LLM tell us about cities?}

\usepackage{bibentry}

\author{
Zhuoheng Li\textsuperscript{\rm 1},
Yaochen Wang\textsuperscript{\rm 1},
Zhixue Song\textsuperscript{\rm 1},
Yuqi Huang\textsuperscript{\rm 1},
Rui Bao\textsuperscript{\rm 1},
Guanjie Zheng\textsuperscript{\rm 2},
Zhenhui Jessie Li\textsuperscript{\rm 1}
}
\affiliations{
\textsuperscript{\rm 1}Yunqi Academy of Engineering, Hangzhou, Zhejiang, China\\
\textsuperscript{\rm 2}Shanghai Jiaotong University, Shanghai, Shanghai, China\\
zhlii@umich.edu, misakiwang74@gmail.com, kmno4.zhixue@gmail.com,\\
hhhuangq@sjtu.edu.cn, 851756936@sjtu.edu.cn, gjzheng@sjtu.edu.cn,\\
jessielzh@gmail.com
}

\begin{document}
\maketitle

\begin{abstract}







This study explores the capabilities of large language models (LLMs) in providing knowledge about cities and regions on a global scale. We employ two methods: directly querying the LLM for target variable values and extracting explicit and implicit features from the LLM correlated with the target variable. Our experiments reveal that LLMs embed a broad but varying degree of knowledge across global cities, with ML models trained on LLM-derived features consistently leading to improved predictive accuracy. Additionally, we observe that LLMs demonstrate a certain level of knowledge across global cities on all continents, but it is evident when they lack knowledge, as they tend to generate generic or random outputs for unfamiliar tasks. These findings suggest that LLMs can offer new opportunities for data-driven decision-making in the study of cities.


\end{abstract}

\section{Introduction}

Cities are of significant importance, as more than half of the world's population now resides in cities. 
With the availability of rich city data, LLMs inherently encode extensive information about various aspects of our cities. 
LLMs could open new possibilities for city research, as illustrated in the following examples.

\begin{figure}[thb]
    \centering
    \includegraphics[width=0.48\textwidth]{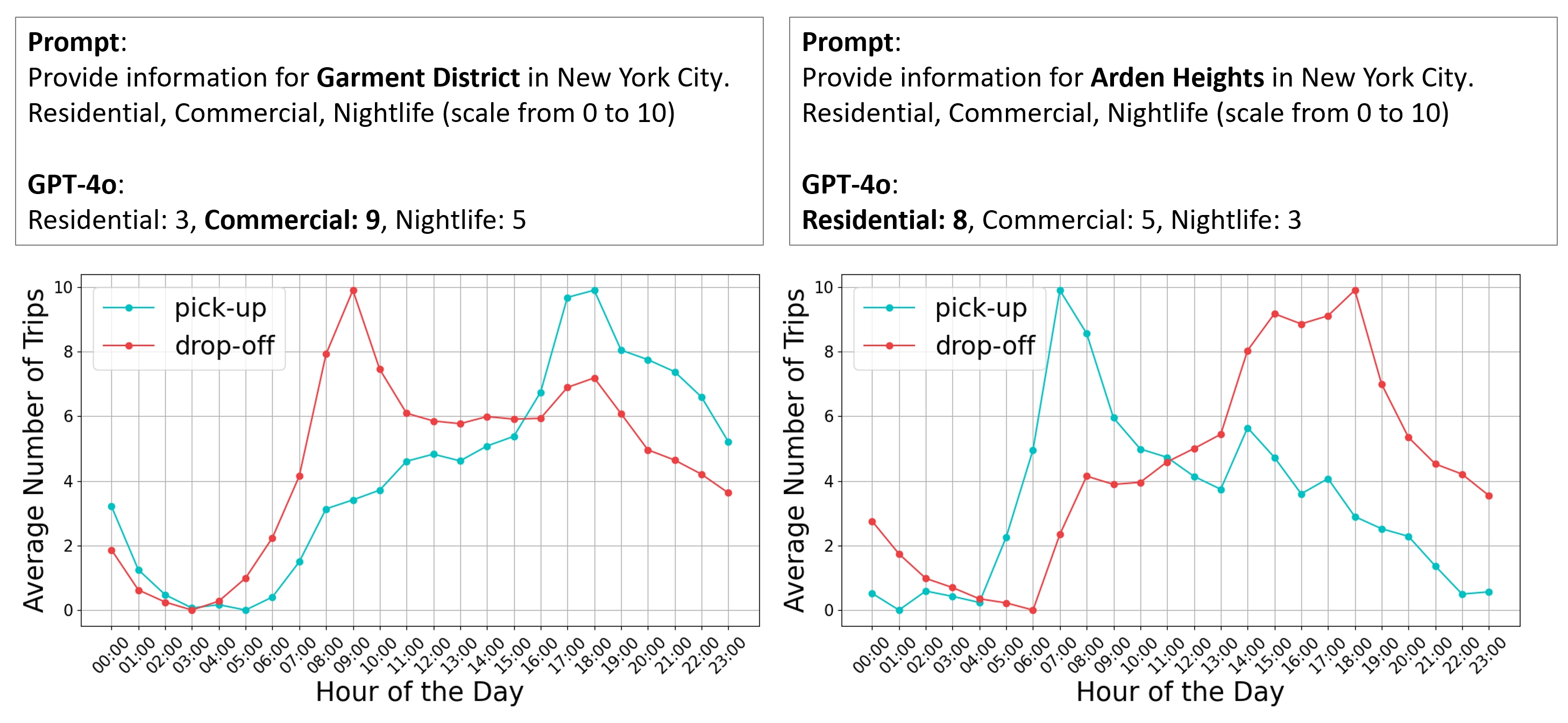}
    \caption{Taxi pick-up and drop-off patterns in NYC: the commercial Garment District (left) shows morning drop-off and evening pick-up peaks, while the residential Arden Heights (right) shows opposite trends.}
    \label{fig:intro}
\end{figure}

\medskip 
\noindent \textbf{Example 1. Region-level knowledge inside cities.}
Using NYC open data~\cite{nyctlctripdata}, Figure~\ref{fig:intro} shows the 24-hour taxi pick-up and drop-off count for two zones in New York City (NYC). Since the 24-hour data are aggregated from raw pick-up and drop-off records, LLMs naturally do not have access to this specific information. When directly asked to generate a 24-hour curve, an LLM will either acknowledge that it lacks this data or produce random numbers. However, LLMs can provide insights into the functions of these zones within NYC, and the semantics of these functions often align well with our understanding of traffic patterns. For example, the LLM identifies the Garment District primarily as a commercial zone, exhibiting a drop-off peak in the morning and a pick-up peak in the evening. In contrast, the LLM classifies Arden Heights as a predominantly residential area, showing the opposite pattern. These functional characteristics can be leveraged as features to model traffic.

This example demonstrates that although LLMs may not have direct access to the target variable, they can identify features that could be used to model it. This implication is significant. If an LLM can estimate features for \emph{all} cities and the correlation between these features and traffic patterns can be generalized,  we could potentially derive traffic patterns for any city, even those that do not release traffic data as comprehensively as NYC does. 

\begin{figure}[thb]
    \centering
    \includegraphics[width=0.48\textwidth]{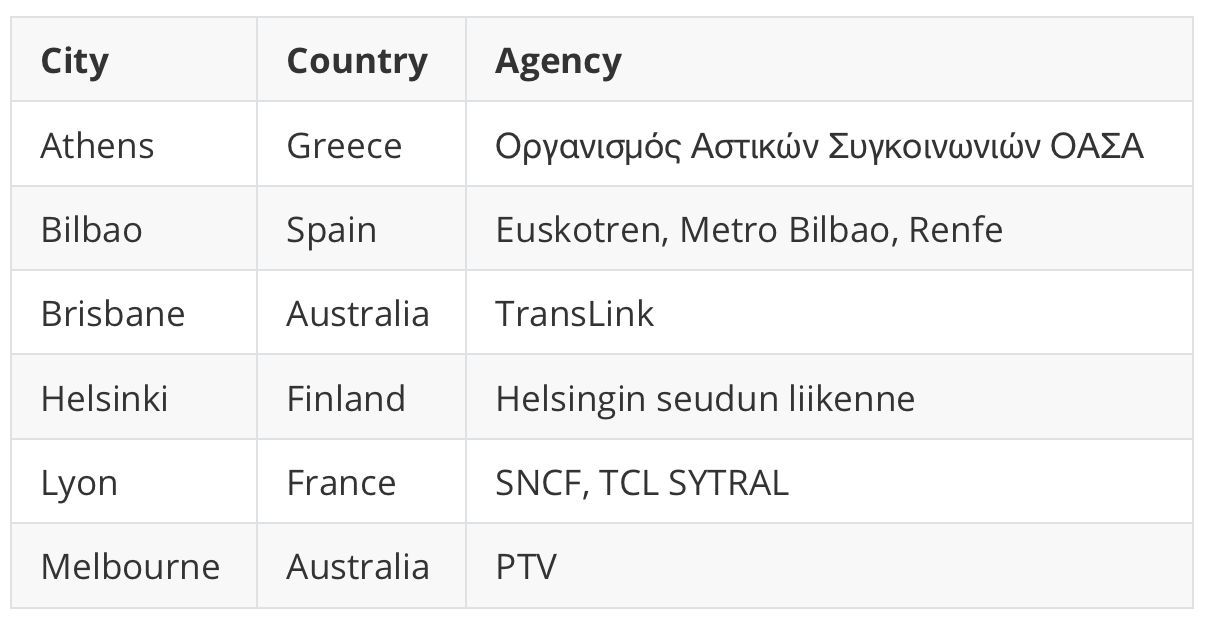}
    \caption{Public transportation data are collected from various agencies. Source: \cite{verbavatz2020access}.}
    \label{fig:intro-ex2}
\end{figure}

\medskip 
\noindent \textbf{Example 2. City-level knowledge across the globe. } Public transportation is a core topic in city research. However, since different cities around the world release data in various formats, collecting public transportation data is a tremendous effort. For instance, a recent study~\cite{verbavatz2020access} compiles public transportation data from 85 cities in OECD (Organization for Economic Cooperation and Development) countries, spanning across 25 nations.
Figure~\ref{fig:intro-ex2} shows a snippet of the data sources. As seen, these countries use different languages, and some data sources include terms that may be well-known locally but unfamiliar to outsiders (e.g., SNCF, which stands for Société nationale des chemins de fer français, is France's national state-owned railway company). Even within the same country, data sources can come from different agencies (e.g., PTV of Melbourne and TransLink of Brisbane, both in Australia). This example illustrates the significant challenges involved in collecting city data across global cities.

\begin{figure}[h]
    \centering
    \includegraphics[width=0.45\textwidth]{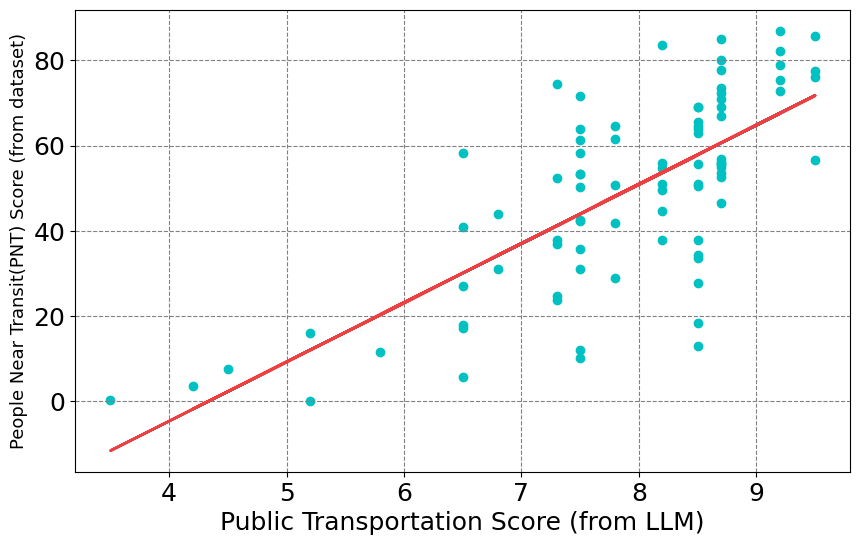}
    \caption{A strong correlation between public transportation scores retrieved from LLM and the public transportation metric computed from raw transportation data~\cite{verbavatz2020access}.}
    \label{fig:pnt}
\end{figure}

Meanwhile, LLMs possess a broad understanding of cities across the globe. As shown in Figure~\ref{fig:pnt}, when asked to score public transportation for these 85 cities, the LLM’s responses strongly correlate with the public transportation metric, People Near Transit (PNT), calculated in the dataset~\cite{verbavatz2020access}. We do not intend to suggest that the effort of collecting raw data should be replaced by LLMs. However, we want to emphasize that LLMs can easily scale up studies to include global cities with minimal effort. In situations where no data is available, having some insight from an LLM is better than having none, especially when attempting to generalize a scientific study to a global scale. While we do not believe that data from LLMs should be directly trusted to derive rigorous scientific findings, we do think that this data can be valuable in helping to test some hypotheses.

\medskip
Motivated by the examples above, we conduct extensive experiments to explore what LLMs know about cities. Our experiments cover more than 40 tasks, each involving hundreds of cities on average. These tasks encompass various aspects of cities, including the environment, transportation, energy, crime, industry, and more. We also experiment with different ways to prompt the LLM and test various models. Through these experiments, we discover many interesting findings, with our key insight being that LLMs know something about everywhere.

In summary, our contributions are as follows:
\begin{itemize}
    \item We systematically investigate the knowledge in LLMs about cities, providing a comprehensive analysis across a broad range of city-related tasks.
    \item We demonstrate that LLMs can effectively identify and leverage latent features within both cities and regions, even in the absence of direct data, offering a novel approach to modeling urban and regional phenomena and generalizing insights across different areas. This suggests a scalable method for extending research to locations with limited or no data availability.
    \item We provide valuable experimental results that pave the way for future studies on the application of LLMs in city research, opening new possibilities for understanding global cities.
\end{itemize}

\section{Related Work}

\subsection{LLM as a Knowledge Base}

By pre-training large language models on vast amounts of data, they acquire a significant amount of knowledge. Since the advent of pre-trained language models, numerous studies have sought to quantify the extent of knowledge contained within these models~\cite{roberts2020much}. A common focus has been on evaluating how well these models respond to factual knowledge~\cite{petroni2019language, jiang2020can}. Other areas of investigation include commonsense knowledge~\cite{davison2019commonsense}, time-sensitive knowledge~\cite{dhingra2022time}, biomedical knowledge~\cite{sung2021can}, geospatial knowledge~\cite{manvi2023geollm}, among others. Our work draws inspiration from GeoLLM~\cite{manvi2023geollm}, which explores using LLMs to infer global geospatial knowledge such as population and education. However, we specifically focus on cities, examining how to extract city knowledge and exploring how this knowledge correlates with open city data.














\subsection{Extracting Knowledge from LLM}
LLMs are known to embed extensive knowledge, yet the challenge lies in effectively extracting this knowledge in a desired way. A common approach involves \emph{fine-tuning} the LLM by adapting its pre-trained weights to a new data domain, thereby enabling its prediction capabilities in that area~\cite{radford2018improving}. Recent years have seen the development of efficient fine-tuning frameworks like LoRA~\cite{hulora}, which optimizes the process by using low-rank matrices to approximate necessary adjustments instead of modifying entire weight matrices. Another group of research focuses on \emph{prompt tuning}~\cite{zhong2021factual}, which seeks to design prompts that effectively leverage the pre-trained model’s knowledge in new domains. Crafting effective prompts is notoriously challenging~\cite{qin2021learning}; thus, strategies like chain-of-thought reasoning~\cite{wei2022chain} and instruction templates~\cite{longpre2023flan} have demonstrated more effective in guiding LLMs toward accurate responses. Rather than adapting the pre-trained model to a specific new field, we propose to investigate the extent of city knowledge that LLMs can reveal across various aspects. Our goal is to demonstrate that LLMs can achieve comparable accuracy to traditional feature engineering approaches but with significantly less effort.

\subsection{LLM for City Tasks}

Recent studies have been utilizing the large language model (LLM) to better tackle traffic-related tasks~\cite{jin2024position,minaee2024large}. Generally, the utilization of LLM can be divided into three categories. 
(1) \emph{LLM as prediction model.} Most of the current methods aim to use LLM as a predictor. Hence, the critical step is to convert the time series data into the space of LLM, by quantizing the time series into discrete tokens~\cite{ansari2024chronos} through methods like VQ-VAE or training a separate encoder to align the time series to LLM semantic space~\cite{zhou2023one}. Thus, LLM can be used to make time series predictions~\cite{gruver2024large} like traffic prediction.  
(2) \emph{Fine-tuned LLM as a prediction model. } Recent studies further explored utilizing the traffic time series data to fine-tune the pre-trained LLM model~\cite{jintime,cao2023tempo,chang2023llm4ts} to achieve better prediction results. Extra background information like date and hour can be extracted from the time series to help LLM better fine-tune~\cite{guo2024towards}. 
(3) \emph{LLM to enhance feature.} Researchers also try to use LLM as a feature extractor~\cite{xue2023promptcast} to reveal more information from the time series data and use this information to further make predictions. (4) \emph{LLM to interpret instructions to run traffic models.} Some studies also use the power of LLM to facilitate a front-end chatbot in which people can call several traffic analysis tools~\cite{zhang2024trafficgpt}. In contrast, we are working on a completely different problem, in which we aim to use LLM to extract external long-tail knowledge and verify its effectiveness on various tasks, instead of one specific prediction task. 

\section{Problem and Method}

Given a set of cities (or regions in a city) \( C = \{c_1, c_2, \dots, c_n\} \), we are interested in obtaining the target variable values for all cities (or regions) in \( C \). Let the target variable be denoted by \( y \). The goal is to determine the values \( y(c_i) \) for each city \( c_i \) in the set \( C \). Cities or regions in the cities are treated the same in the method. We will use cities for simplicity in the description of the method.

We conduct experiments in two categories:
\begin{itemize}
    \item Directly asking LLM. This method is to directly ask LLM about the values of the target value for the city of interest.
    \item Extract features and train an ML model. This line of methods is to ask LLM about the features that potentially correlate with the target variable for the city of interest. Using these features retrieved for each city, we then train a machine learning model to predict the target variables across the set of cities.
\end{itemize}

We will describe the methods in detail in the rest of the section. 


\subsection{Directly Asking LLM}
When directly asking LLM, the prompt includes both the city name and the target variable. The following is a template applicable to all tasks.

\begin{tcolorbox}
\tiny
\begin{verbatim}
Your task is to provide target information about {Place}.

Organize your answer in a JSON object containing the 
following keys:
- zone: The name of the city, {Place}
- pred: {Target} {Unit} in {Place} 
\end{verbatim}
\end{tcolorbox}

 If the city or the target variable changes, we simply adjust the ``Target'', ``Place'', and ``Unit'' accordingly.



\subsection{Implicit Feature Extraction}

To extract the features for the target variable, we obtain the last hidden states from the LLM and then treat them as features. These features may not be human-understandable but they encapsulate knowledge relevant to the city and the target variable. 

We use the same prompt as the one used in directly asking LLM. The last hidden layer is used to extract features. However, directly using the last layer is problematic due to its size. For example, the last hidden layer of Llama3.1-8B is the size of $37 \times 4096$. Since our data samples are usually in the scale of hundreds, the high dimensionality will cause the under-fitting of the machine learning models. It is necessary to conduct dimension reduction for the raw features from the last hidden layer.

\begin{figure}[thb]
    \centering
    \includegraphics[width=0.48\textwidth]{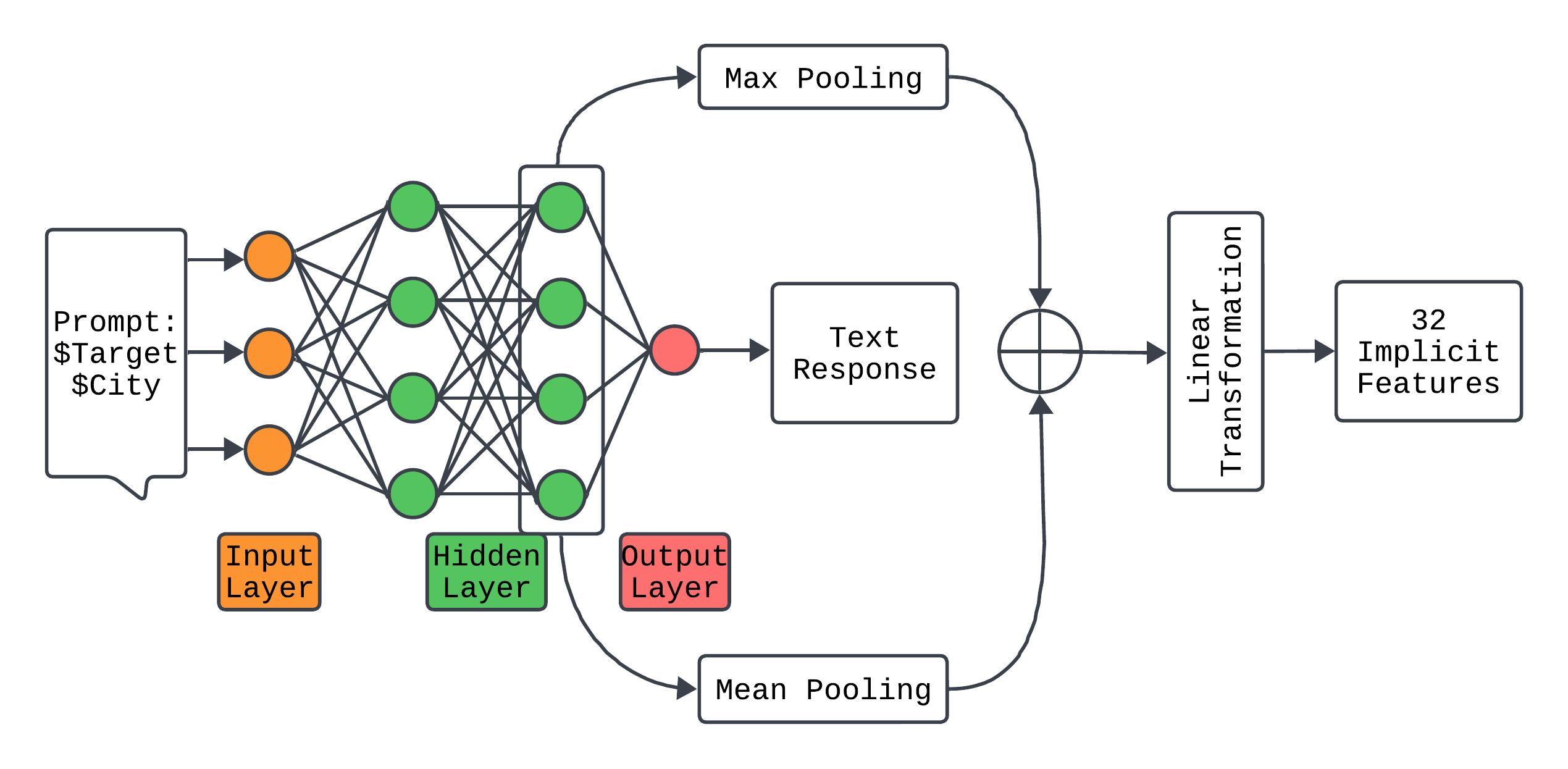}
    \caption{Features of a specific task are extracted from the last hidden state using Llama3.1-8B. }
    \label{fig:extract_target}
\end{figure}

As shown in Figure~\ref{fig:extract_target}, to conduct dimension reduction, we first use a concatenated mean-max pooling technique, followed by a linear transformation, inspired by common practices in NLP~\cite{Lin2017} and computer vision~\cite{Thomas2016}. The dimension of features is 32 by default in experiments. A machine learning model is further trained on the extracted features and the target variable. 


We could only conduct this experiment on the open-source LLMs. LLMs such as chatGPT only provide API access, limiting the access of the hidden layers. In the experiments, we use Llama3.1-8B-Instruct by default. We also conduct experiments using Llama3.1-70B-Instruct. The difference between these two models is marginal in our tasks.


\subsection{Explicit Feature Extraction}

Another method to extract features is to explicitly include the feature names in the prompt. Different from implicit feature extraction, such an explicit approach can extract features that are human-understandable. The following example is a template applicable to all tasks. 


\begin{tcolorbox}
\tiny
\begin{verbatim}
Your task is to provide feature information about {Place}. 

Please provide information about <{feature1}>, 
<{feature2}>, and <{feature3}> in {Place}:

{feature1}: {description1}
{feature2}: {description2}
{feature3}: {description3}

Please organize your answer in a JSON object containing 
the following keys:
"{feature1}" (scale from 0.0 to 10.0),
"{feature2}" (scale from 0.0 to 10.0),
"{feature3}" (scale from 0.0 to 10.0)
\end{verbatim}
\end{tcolorbox}


Though explicit features are good for human understanding, it is tricky to design what features to be asked. Especially for the unfamiliar target variable, we may have little knowledge of which features might be relevant. Even for the target variable that we are familiar with, we may not be sure which wording we should use for features.

\begin{figure}[h]
\vspace{-3pt}
    \centering
    \includegraphics[width=0.48\textwidth]{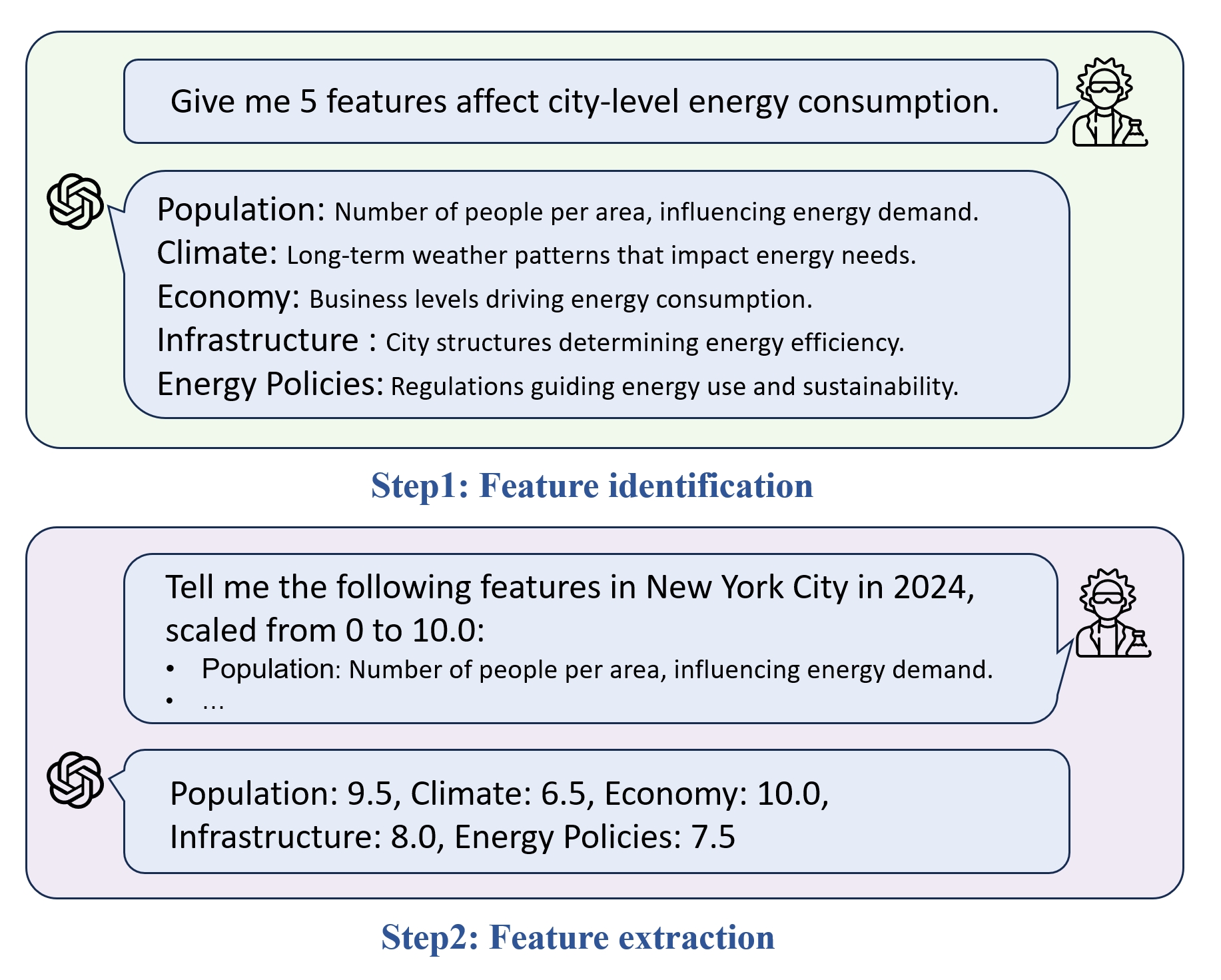}
    \caption{Extract relevant features impacting city-level energy consumption and provide their scaled values for a specific location (e.g. New York City in 2024) to better understand the factors influencing energy usage.} 
    \label{fig:extract_features}
\vspace{-3pt}
\end{figure}

We find that instead of handcrafting features by humans, it is more effective to ask LLM about the features to use. LLM not only serves as feature extraction but also feature identification in this method. We design an interactive feature extraction framework as shown in Figure~\ref{fig:extract_features}. In feature identification step, LLM provides the general knowledge related to the target variable. In feature extraction step, LLM provides specific knowledge about certain features for the city of interest. In this framework, the feature extraction is more coherent using the knowledge and the wording that LLM inherently encodes. The correlation with the target variable is often more significant for LLM-extracted features, leading to better model performance compared to manually selected features.

\begin{table*}[]
    \centering
    \renewcommand{\arraystretch}{1.4}
    \resizebox{1.0\linewidth}{!}{
    \begin{tabular}{l|l|l|l|l|rr|rrr|r}
        \toprule
        \multicolumn{2}{l|}{} &  & & & \multicolumn{2}{c|}{\textbf{GPT-4o}} & \multicolumn{3}{c|}{\textbf{Llama3.1-8B}} & \\ \cline{6-10}
        \multicolumn{2}{l|}{} & \textbf{Task} & & & \textbf{Explicit} & \textbf{Direct} & \textbf{Explicit} & \textbf{Implicit} & \textbf{Direct} & \textbf{No} \\ 
        \multicolumn{2}{l|}{\multirow{-3}{*}{\textbf{Category}}} & \textbf{ID} & \multirow{-3}{*}{\textbf{Data}} & \multirow{-3}{*}{\textbf{Coverage}} & \textbf{Feature} & \textbf{Ask} & \textbf{Feature} & \textbf{Feature} & \textbf{Ask} & \textbf{Feature} \\ \hline
        
        & & 1 & $NO_2$ & 749 cities, Europe & \cellcolor{lgray}{\textbf{3.35}} & 4.53 & \textbf{4.21} & 4.37 & 6.10 & 4.55 \\ \cline{3-11} 
        & & 2 & $O_3$ & 749 cities, Europe & \cellcolor{lgray}{\textbf{1187.05}} & 3995.10 & 1507.91 & \textbf{1468.87} & 4011.17 & 1606.76 \\ \cline{3-11} 
        & & 3 & $PM_{10}$ & 749 cities, Europe & 4.20 & \cellcolor{lgray}{\textbf{3.87}} & 5.00 & \textbf{4.64} & 7.70 & 5.55 \\ \cline{3-11} 
        & & 4 & $PM_{2.5}$ & 749 cities, Europe & 3.59 & \cellcolor{lgray}{\textbf{2.16}} & 3.80 & \textbf{3.51} & 4.58 & 4.13 \\ \cline{3-11} 
        & & 5 & Methane & 347 cities, China & \cellcolor{lgray}{\textbf{24654.91}} & 45339.00 & 34225.74 & \textbf{28534.40} & err & 35258.76 \\ \cline{3-11} 
        & & 6 & Carbon emission & 343 global cities & \cellcolor{lgray}{\textbf{10.26}} & 13.37 & 14.66 & \textbf{13.41} & 353.46 & 12.51 \\ \cline{3-11} 
        & & 7 & Household $CO_2$ & 52 cities, Japan & \cellcolor{lgray}{\textbf{139.98}} & 853.42 & \textbf{164.29} & 181.92 & 4192.01 & 170.19 \\ \cline{3-11} 
        & & 8 & Total $CO_2$ & 82 entities, Russia & \cellcolor{lgray}{\textbf{23.67}} & 23.96 & 25.24 & 28.64 & \textbf{24.55} & 25.61 \\ \cline{3-11} 
        & \multirow{-9}{*}{\rotatebox{90}{\textbf{Environment}}} & 9 & Water withdrawal & 391 cities, China & \cellcolor{lgray}{\textbf{13.40}} & 17.38 & \textbf{14.89} & 15.57 & err & 15.78 \\ \cline{2-11} 
        
        & & 10 & Total energy & 331 cities, China & \cellcolor{lgray}{\textbf{886.75}} & 1208.01 & 1414.11 & \textbf{1400.86} & err & 1250.45 \\ \cline{3-11} 
        & & 11 & Traditional energy & 331 cities, China & \cellcolor{lgray}{\textbf{851.10}} & 2250.13 & \textbf{1202.83} & 1293.12 & err & 1154.61 \\ \cline{3-11} 
        & \multirow{-3}{*}{\rotatebox{90}{\textbf{Energy}}} & 12 & Renewable energy & 331 cities, China & \cellcolor{lgray}{\textbf{110.72}} & 2934.21 & \textbf{133.37} & 135.49 & err & 143.64 \\ \cline{2-11} 
        
        & & 13 & Dengue & 1541 global cities & \cellcolor{lgray}{\textbf{1858.82}} & 3309.80 & 1945.33 & \textbf{1896.27} & err & 1947.40 \\ \cline{3-11} 
        & & 14 & COVID-19 & 415 counties, USA & 2234.81 & \cellcolor{lgray}{\textbf{2159.20}} & 2508.96 & \textbf{2367.35} & 2782.53 & 2784.65 \\ \cline{3-11} 
        & & 15 & Life expectancy & 29 cities, England & \cellcolor{lgray}{\textbf{2.59}} & 13.10 & 2.81 & \textbf{2.61} & 10.70 & 2.90 \\ \cline{3-11} 
        & \multirow{-4}{*}{\rotatebox{90}{\textbf{Health}}} & 16 & Health access & 374 districts, UK & \cellcolor{lgray}{\textbf{5.38}} & 16.39 & 7.33 & \textbf{7.02} & 20.21 & 6.19 \\ \cline{2-11} 
        
        & & 17 & Avg travel time & 387 global cities & \cellcolor{lgray}{\textbf{205.52}} & 370.43 & 246.01 & \textbf{228.39} & 1089.14 & 269.47 \\ \cline{3-11} 
        & & 18 & Avg speed & 387 global cities & \cellcolor{lgray}{\textbf{7.05}} & 10.44 & 8.75 & \textbf{8.10} & 16.42 & 10.02 \\ \cline{3-11} 
        & & 19 & PNT (500m) & 85 global cities & \cellcolor{lgray}{\textbf{10.54}} & 14.27 & 13.49 & \textbf{11.09} & 26.83 & 14.16 \\ \cline{3-11} 
        & & 20 & PNT (1000m) & 85 global cities & \cellcolor{lgray}{\textbf{13.22}} & 19.41 & 18.77 & \textbf{15.53} & 23.76 & 20.19 \\ \cline{3-11} 
        & \multirow{-5}{*}{\rotatebox{90}{\textbf{Transport}}} & 21 & PNT (1500m) & 85 global cities & \cellcolor{lgray}{\textbf{13.98}} & 19.92 & 20.06 & \textbf{17.94} & 28.34 & 22.55 \\ \cline{2-11} 
        
        & & 22 & Total industry & 245 cities, China & \cellcolor{lgray}{\textbf{3357.65}} & 3508.40 & \textbf{3914.05} & 5040.15 & err & 5296.17 \\ \cline{3-11} 
        & & 23 & Mining & 245 cities, China & \cellcolor{lgray}{\textbf{314.64}} & 329.82 & \textbf{324.00} & 344.96 & err & 324.14 \\ \cline{3-11} 
        & & 24 & Manufacture & 245 cities, China & \cellcolor{lgray}{\textbf{3224.63}} & 3383.90 & \textbf{3805.45} & 4735.16 & err & 5010.07 \\ \cline{3-11} 
        & \multirow{-4}{*}{\rotatebox{90}{\textbf{Industry}}} & 25 & Utilities & 245 cities, China & \textbf{309.55} & 372.22 & \cellcolor{lgray}{\textbf{295.15}} & 314.23 & err & 378.27 \\ \cline{2-11} 
        
        & & 26 & Patent & 487 global cities & \cellcolor{lgray}{\textbf{2490.46}} & 3305.90 & \textbf{2793.08} & 2851.43 & err & 2826.34 \\ \cline{3-11} 
        & & 27 & Avg home value & 200 cities, USA & 184.32 & \cellcolor{lgray}{\textbf{23.96}} & \textbf{202.87} & 223.66 & 292.84 & 238.02 \\ \cline{3-11} 
        \multirow{-28}{*}{\rotatebox{90}{\textbf{City-Level}}} & \multirow{-3}{*}{\rotatebox{90}{\textbf{Develop}}} & 28 & Material stocks & 337 cities, China & \cellcolor{lgray}{\textbf{33.71}} & 106.68 & 34.68 & \textbf{34.24} & err & 36.60 \\ \hline
        
        & & 29 & NYC pick-up & 256 regions & \textbf{1.47} & 3.11 & \cellcolor{lgray}{\textbf{1.45}} & 1.61 & 3.23 & 1.65 \\ \cline{3-11} 
        & & 30 & NYC drop-off & 256 regions & \cellcolor{lgray}{\textbf{1.69}} & 3.95 & \textbf{1.72} & 1.78 & 3.61 & 1.79 \\ \cline{3-11} 
        & & 31 & Chicago pick-up & 167 regions & \cellcolor{lgray}{\textbf{1.82}} & 3.19 & \textbf{1.91} & 1.96 & 3.17 & 1.96 \\ \cline{3-11} 
        & & 32 & Chicago drop-off & 167 regions & \cellcolor{lgray}{\textbf{1.75}} & 3.45 & 2.07 & \textbf{1.96} & 3.48 & 2.02 \\ \cline{3-11} 
        & & 33 & DC pick-up & 129 regions & \cellcolor{lgray}{\textbf{1.89}} & 3.02 & \textbf{1.91} & 1.95 & 4.48 & 1.97 \\ \cline{3-11} 
        & \multirow{-6}{*}{\rotatebox{90}{\textbf{Transport}}} & 34 & DC drop-off & 129 regions & \cellcolor{lgray}{\textbf{1.85}} & 3.20 & \textbf{1.87} & 1.90 & 4.83 & 1.89 \\ \cline{2-11} 
        
        & & 35 & NYC crime & 177 regions & \cellcolor{lgray}{\textbf{1.85}} & 3.47 & 2.35 & \textbf{2.16} & 2.87 & 1.90 \\ \cline{3-11} 
        & & 36 & Chicago crime & 77 regions & \cellcolor{lgray}{\textbf{1.30}} & 3.15 & \textbf{1.99} & 2.26 & 3.03 & 2.27 \\ \cline{3-11} 
        & & 37 & DC crime & 41 regions & \cellcolor{lgray}{\textbf{1.90}} & 5.19 & 2.87 & \cellcolor{lgray}{\textbf{1.90}} & 4.01 & 2.35 \\ \cline{3-11} 
        & \multirow{-4}{*}{\rotatebox{90}{\textbf{Crime}}} & 38 & Cincinnati crime & 67 regions & \cellcolor{lgray}{\textbf{1.60}} & 4.70 & 1.92 & \textbf{1.79} & 3.57 & 1.74 \\ \cline{2-11} 
        
        & & 39 & NYC 311 & 177 regions & \cellcolor{lgray}{\textbf{1.84}} & 4.43 & 2.15 & \textbf{1.85} & 3.59 & 2.19 \\ \cline{3-11} 
        & & 40 & Austin 311 & 69 regions & \textbf{1.75} & 4.19 & 2.72 & \cellcolor{lgray}{\textbf{1.67}} & 3.33 & 2.57 \\ \cline{3-11} 
        \multirow{-13}{*}{\rotatebox{90}{\textbf{Region-Level}}} & \multirow{-3}{*}{\rotatebox{90}{\textbf{311}}} & 41 & Cincinnati 311 & 47 regions & \cellcolor{lgray}{\textbf{2.09}} & 5.03 & 3.22 & \textbf{2.16} & 3.25 & 2.78 \\ \bottomrule
    \end{tabular}
    }

\caption{Results on City-level and Region-level tasks for cities all over the world. The values represent RMSE (Root Mean Square Error) compared to ground truth. ``err" indicates the return of answers may be entirely unrelated to the question. Units and notations are detailed in the supplementary material. }
    \label{tab:overall-results}
\end{table*}

\section{Experiment}

\subsection{Experiment Settings}

We conduct experiments about 41 various tasks in 8 domains on two levels of data, region-level (each data point represents a census block, community area, or district) and city-level (each data point represents a city). For evaluation, we use the Root Mean Square Error (RMSE) as the primary metric. 

The compared methods are listed below.
\begin{itemize}
    \item \textbf{Exp-Feature}: Given a task, explicitly ask LLM for related feature names and then ask LLM for feature values.
    \item \textbf{Imp-Feature}: Ask LLM about the target variable and implicitly obtain the last hidden layer vector to represent the city/region.
    \item \textbf{Direct-Ask}: Directly ask LLM for predictions on the target variable.
    \item \textbf{No-Feature}: Use the average value of the target as the predicted value.
\end{itemize}

Note that, the Exp-Feature and Imp-Feature are further fed into an ML model. A set of frequently used ML models are tried including Decision Tree, Random Forest, Gradient Boosting, XGBoost, AdaBoost, and Linear Regressor. The best results are reported. We use 5-fold cross-validation in the ML setting.

Regarding computational resources, we use machines equipped with either four NVIDIA A30 GPUs or eight NVIDIA A100 GPUs. For the model, note that all references to Llama 3.1-8B in this paper refer specifically to Llama 3.1-8B-Instruct. Llama3.1-8B is locally deployed on A30 GPUs, while GPT-4o is accessed via API. Due to the limitations of the GPT-4o API, which does not allow access to hidden layers, the Imp-Feature method was implemented solely using the locally deployed Llama3.1-8B to extract the necessary hidden layers.

To illustrate, on the Carbon emission dataset~\cite{nangini_global_2019}, which includes 343 global cities, the Exp-Feature and Direct-Ask experiments on GPT-4o each take approximately 50 minutes. Similarly, on Llama3.1-8B, these experiments also take around 50 minutes, while the Imp-Feature experiment requires only about 1 minute. The No-Feature experiment, which does not involve GPU computation, takes only a very short amount of time.




\subsection{Overall Results}

Table~\ref{tab:overall-results} shows the comprehensive results of 41 tasks. The key findings are as follows: 

\begin{itemize}
    \item LLMs generally provide meaningful responses, enabling us to predict target values effectively, whether through Direct-Ask or feature-based methods. For popular datasets (i.e., datasets that might be public and easy-to-be-browsed online), Direct-Ask yields the smallest error. This means \textit{LLM directly knows the answer} (like the $PM_{2.5}$, $PM_{10}$, COVID-19, and Home value cases). 
    \item For most of the cases, extracting features improves the performance over Direct-Ask and No-Feature. This means that even for target values that LLM does not directly know, \textit{LLM can help to extract useful features to assist the prediction}.
    \item  In terms of different LLM models, \textit{GPT-4o performs better in general}. This selection of various LLMs will be further illustrated later in the experiment. 
    \item It is interesting to see that Exp-Feature and Imp-Feature using Llama3.1-8B show a tie in their performance, indicating that \emph{explicitly extracting features is recommended in practice} because not all LLMs provide the hidden layers as Llama does. 
\end{itemize}


\subsection{Q1: Can we determine when LLM knows or does not know the answer?}

Although it can be fairly difficult to prove what LLM says is correct (without knowing the ground truth), there are some obvious flags when LLM does not know. 

\subsubsection{Detecting consistent generic values.}  
\begin{figure}[htb]
    \centering
    \includegraphics[width=0.45\textwidth]{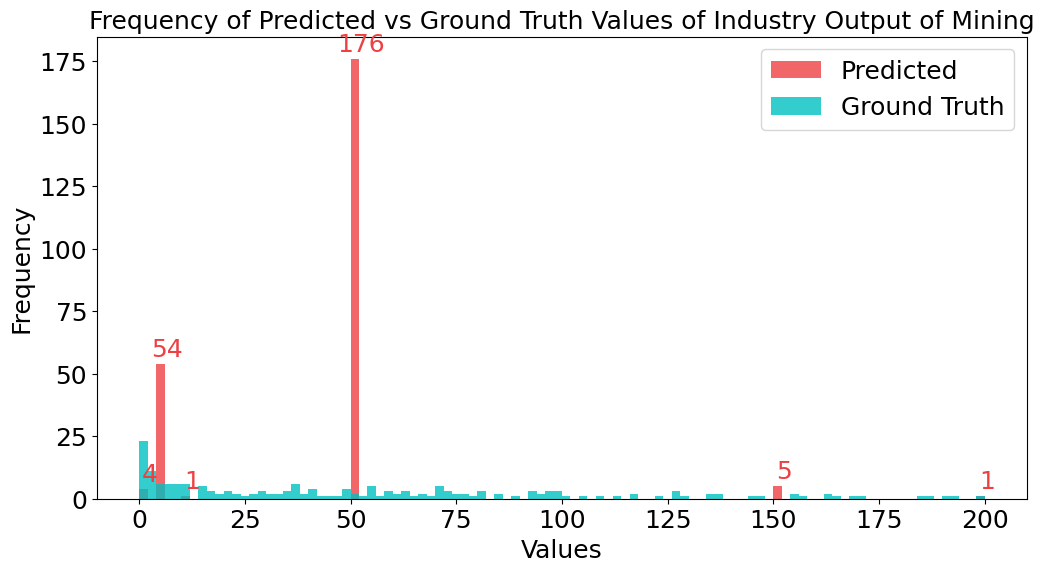}
    \caption{Query results on Mining for 245 cities. 93.9\% queried values are `5' and `50'. This indicates LLM frequently defaults to placeholder values for unknown tasks.}
    \label{fig:VER-I}
\end{figure}

If the LLM frequently produces the same rounded number, such as 50, across various contexts, it suggests that the model may lack specific knowledge about the data in question. This behavior indicates a tendency to default to a generic or placeholder value rather than providing a contextually informed estimate. As shown in Figure~\ref{fig:VER-I}, for the queries on values of Industry Output of Mining for 245 cities in China~\cite{zhang2024water}, LLM outputs certain placeholder values for most of the time (`5' for 54 times, and `50' for 176 times out of the 245 city queries). This is clearly suggesting suspicious generic outputs.

\subsubsection{Detecting inconsistent values among different rounds of generations.} 
\begin{figure}[htb]
    \centering
    \includegraphics[width=0.45\textwidth]{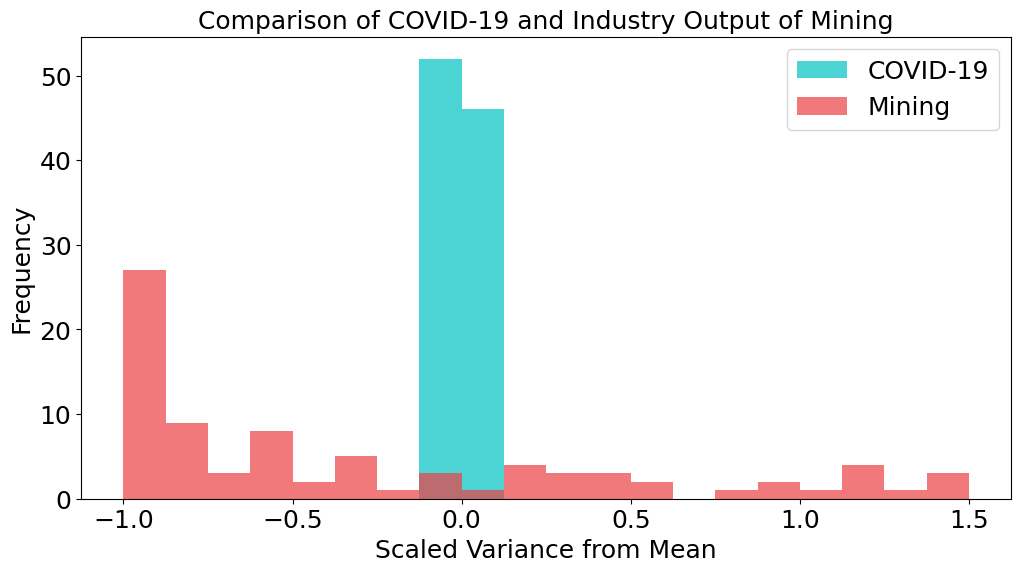}
    \caption{Re-scaled deviation of each sample in 100 queries for two cities in two datasets: COVID-19 (Los Angeles, California) and Mining (Tianjin, China). The responses show much more certainty for COVID-19 compared to Mining, indicating that the LLM is more confident in its knowledge of COVID-19 data but lacks certainty regarding Mining data. }
    \label{fig:VER-II}
\end{figure}

If the outputs of the LLM regarding the same query exhibit large variance, this suggests that the model may be uncertain about the information and is generating responses based on random guesses rather than a solid knowledge of the data. Such inconsistency indicates that the LLM may not be reliably estimating the information, instead oscillating between generic or uncertain predictions. As shown in Figure~\ref{fig:VER-II}, we plot the re-scaled deviation of each sample in 100 queries for two datasets, COVID-19~\cite{cdc_covid_death_counts} and Mining. For a dataset that LLM has high confidence, COVID-19, its output is very stable. While for a dataset that LLM has no clear answer, Mining, its output has a much larger deviation. This aligns with the fact that LLM does a good job in the COVID-19 prediction while a poor job in the Mining prediction. 

\subsection{Q2: Can LLM tell the precise value of the features?}

\begin{figure}[htb]
\centering
\includegraphics[width=0.45\textwidth]{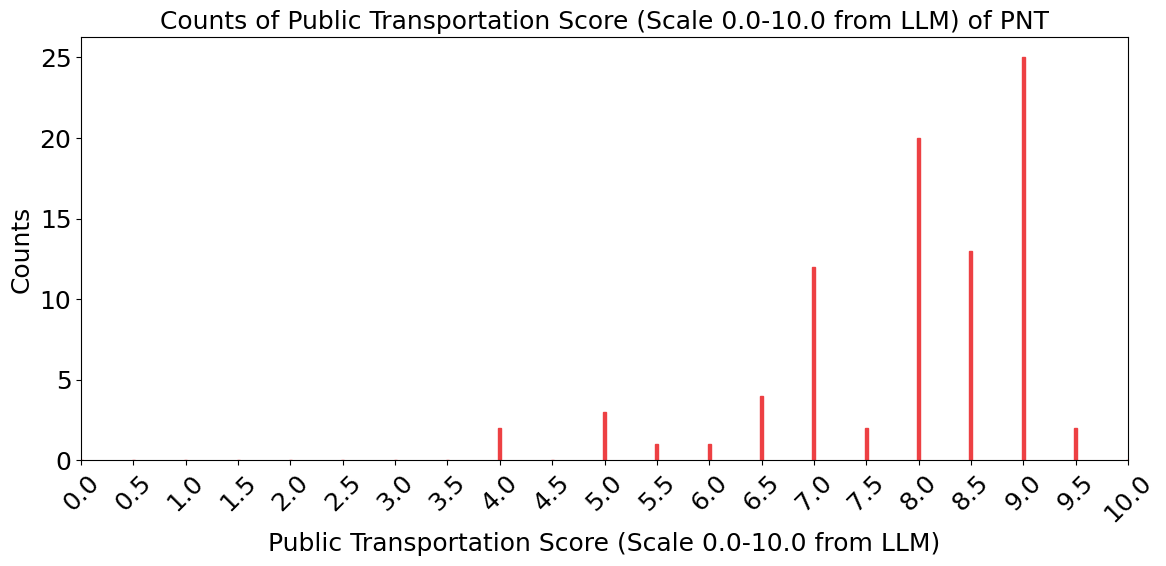} 
    \caption{Count statistics of the public transportation score extracted from LLM using the 0.0-10.0 scale in the PNT dataset. The extracted scores occur at a certain fixed gap value (0.5). }
    \label{fig:fixed_gap}
\end{figure}

When asking LLM for feature values, it is observed that LLM may generate data with some randomness. Here, we ask LLM to generate features for PNT on a scale of 0.0-10.0 and 0.0-100.0. The extracted feature values on a scale of 0.0-10.0 are shown in Figure~\ref{fig:fixed_gap}. A pattern emerges where the generated features tend to cluster around specific intervals (e.g., 5.0, 5.5, 6.0). When changing the feature scale to 0.0-100.0, the original feature values are basically amplified by 10 times. Averaging results across scales of 0.0-10.0 and 0.0-100.0 reveals that over 60\% of standard deviations are less than 1.0, indicating consistent feature extraction by the LLM across different scales, with minor variation added.

\begin{figure}[htb]
\centering
\begin{tabular}{cc}
\includegraphics[width=0.47\textwidth]{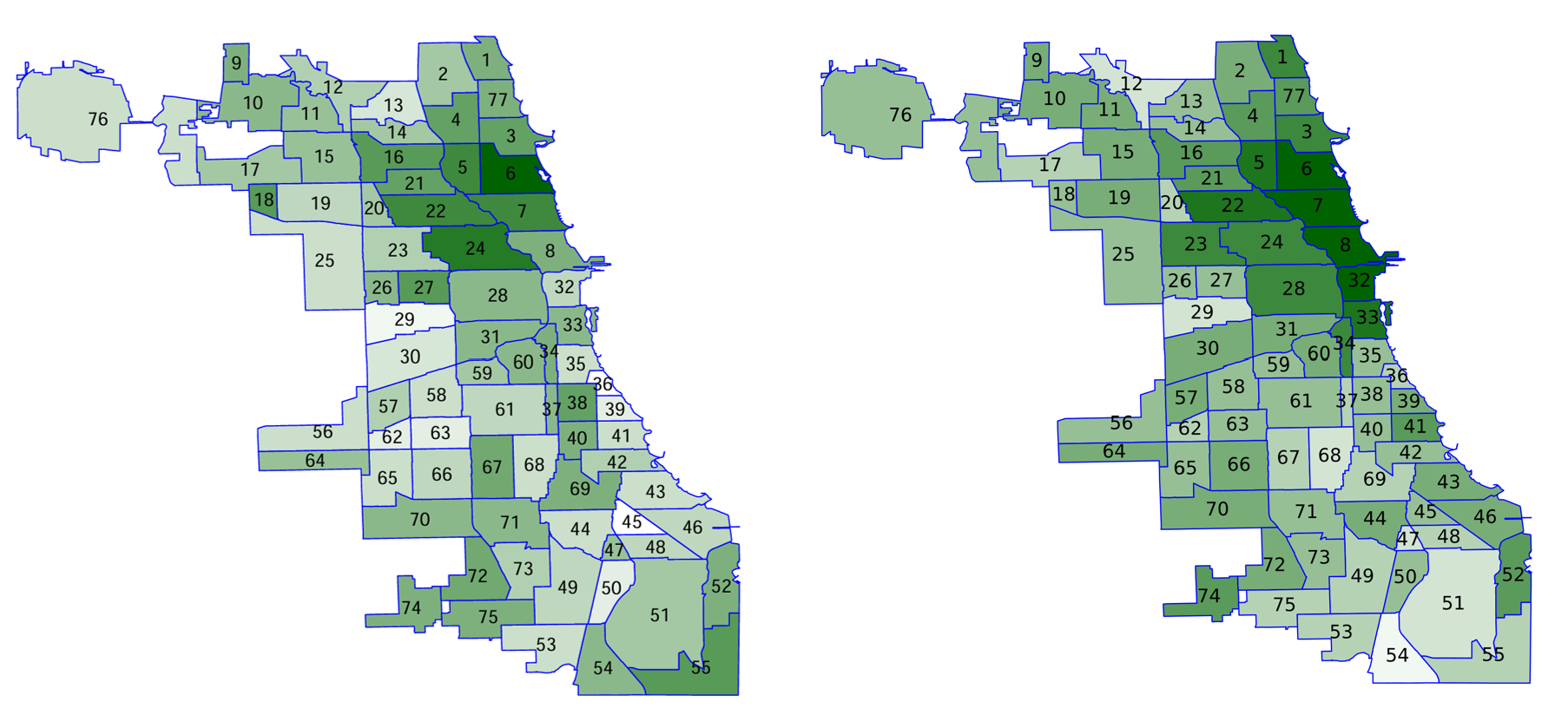}     &  

\end{tabular}
    \caption{Similarity is being observed on Chicago Nightlife Feature (left: POI data processed from \cite{wang2016crime}; right: explicit feature extracted by GPT-4o). Darker colors indicate more active nightlife in the regions.}
    \label{fig:chicago_night}
\end{figure}

Thus, it is easy to conclude that LLM is not outputting the precise value. Since LLM-generated features do help the prediction (as in Table~\ref{tab:overall-results}), it can be inferred that LLM-generated features keep the relative order of samples compared with the ground truth samples. Here, we plot the ground truth nightlife POI data \cite{wang2016crime}  and LLM-generated nightlife POI data for Chicago in Figure~\ref{fig:chicago_night}. We can observe that although the exact values in the two feature sets are different, the relative order among different regions is close. 

\subsection{Q3: Which LLM model performs the best?}

While we have compared GPT-4o and Llama3.1-8B in Table~\ref{tab:overall-results}, here we further compare several frequently-used LLM via API queries. These LLMs are asked to generate the Methane data~\cite{city_methane} of 347 cities in China using the Exp-Feature manner. It is obvious that GPT-4o achieves the best performance and all other three models hold an 8\% - 19\% larger error.

\begin{table}[tbph]
\centering

\renewcommand{\arraystretch}{1.2}
\begin{tabular}{l|c|c}
\toprule
Model                & Methane & +Error \\ \hline
GPT-4o   & \textbf{24654.910} & / \\ \hline
Claude3.5-sonnet & 26610.090    & +7.9\%    \\ \hline
Deepseek-chat & 28310.989  & +14.8\%         \\   \hline
Qwen2-72B-Instruct  & 29364.221  & +19.1\%     \\ 
\bottomrule
\end{tabular}
\caption{Comparison of RMSE across different LLMs on the Methane dataset.  The ``+Error'' column shows the relative percentage increase compared to GPT-4o. }
\label{tab:model_performance}
\end{table}

We further investigate the impact of model size by implicit feature Llama3.1-8B and Llama3.1-70B for the NYC 311 data~\cite{NYC311_2023} three times. The results in Table~\ref{tab:8b70b} show that the model size does not make too much difference.

\begin{table}[htb]
\centering

\renewcommand{\arraystretch}{1.2}
\begin{tabular}{c|ccc|ccc}
\toprule
Model & \multicolumn{3}{c|}{Llama3.1-8B}      & \multicolumn{3}{c}{Llama3.1-70B}         \\ 
Run & 1 & 2 & 3 & 1 & 2 & 3 \\  \midrule
RMSE & 1.81 & 1.89 & 1.97 & 1.89 & 1.98 & 1.91 \\
\bottomrule
\end{tabular}
\caption{Comparison of RMSE results from three runs using Llama3.1 (8B and 70B) on the NYC 311 dataset with Imp-Feature. The findings indicate that a larger model size did not lead to better performance.}
\label{tab:8b70b}
\end{table}


\subsection{Q4: Which language should be used to formulate the prompt?}

We formulate the prompts in Chinese and English respectively and test the Direct-Ask performance on two datasets, Material stocks~\cite{li2023product} and PNT(1500m). Since Material stocks focus on Chinese cities while PNT(1500m) covers worldwide cities, we expect prompts in different languages to exhibit different performances. The results are shown in Table~\ref{tab:language}. For global cities, using English brings a better performance, while for Chinese cities, using Chinese can bring a similar performance as using English.

\begin{table}[htb]
\centering
\renewcommand{\arraystretch}{1.2}
\begin{tabular}{cc|cc}
\toprule
\multicolumn{2}{c}{Material stocks (China)}    &        \multicolumn{2}{|c}{PNT(1500m) (Global)}\\ \cline{1-4} 
English   & Chinese  & English   & Chinese \\ \hline
    {33.706}           & \textbf{33.216} & {\textbf{13.976}} & 15.522             \\ 
    \bottomrule
\end{tabular}
\caption{The impact of Language on Direct-Ask performance across Material Stocks and PNT(1500m), which shows that English performs better for global cities, while Chinese and English are close in performance for Chinese cities.}
\label{tab:language}
\end{table}
\section{Conclusion } 

Our paper conducts extensive experiments to explore what LLM knows about cities. From experiments, there are three key takeaways. (1) \textbf{LLM knows something about everywhere.} Our experiments include global cities on all continents. All the experiments consistently show that LLM either directly knows the answer or contains related features that could be used to train ML models with 18\% improvements over the No-Feature method. 
(2) \textbf{It is obvious to know when LLM does not know.} Most of our tasks are not common knowledge and LLM may have not seen such data before. In these cases, we commonly see that LLM generates generic or random answers. (3) \textbf{Explicit feature extraction is often more effective and versatile}, offering a structured approach to capturing information across both open-source and closed-source models. Though Exp-Feature and Imp-Feature with Llama 3.1-8B perform similarly, Exp-Feature is recommended since not all LLMs provide accessible hidden layers.
Additionally, our method could offer valuable insights into how different models perform in city contexts. 
In conclusion, our work paves the
way for future studies on the application of LLMs in city
research, opening new possibilities for understanding global cities.

\newpage

\bibliography{ref}

\begin{thebibliography}{52}
\providecommand{\natexlab}[1]{#1}

\bibitem[{Ansari et~al.(2024)Ansari, Stella, Turkmen, Zhang, Mercado, Shen, Shchur, Rangapuram, Arango, Kapoor et~al.}]{ansari2024chronos}
Ansari, A.~F.; Stella, L.; Turkmen, C.; Zhang, X.; Mercado, P.; Shen, H.; Shchur, O.; Rangapuram, S.~S.; Arango, S.~P.; Kapoor, S.; et~al. 2024.
\newblock Chronos: Learning the language of time series.
\newblock \emph{arXiv preprint arXiv:2403.07815}.

\bibitem[{Cao et~al.(2023)Cao, Jia, Arik, Pfister, Zheng, Ye, and Liu}]{cao2023tempo}
Cao, D.; Jia, F.; Arik, S.~O.; Pfister, T.; Zheng, Y.; Ye, W.; and Liu, Y. 2023.
\newblock Tempo: Prompt-based generative pre-trained transformer for time series forecasting.
\newblock \emph{arXiv preprint arXiv:2310.04948}.

\bibitem[{CDC(2023)}]{cdc_covid_death_counts}
CDC. 2023.
\newblock Provisional COVID-19 Death Counts in the United States by County.
\newblock Data retrieved from CDC data catalog.
\newblock Effective June 28, 2023, this dataset will no longer be updated. Similar data are accessible from CDC WONDER (https://wonder.cdc.gov/mcd-icd10-provisional.html).

\bibitem[{Chang, Peng, and Chen(2023)}]{chang2023llm4ts}
Chang, C.; Peng, W.-C.; and Chen, T.-F. 2023.
\newblock Llm4ts: Two-stage fine-tuning for time-series forecasting with pre-trained llms.
\newblock \emph{arXiv preprint arXiv:2308.08469}.

\bibitem[{{Chicago Police Department}(2024)}]{chicagocrimes2014}
{Chicago Police Department}. 2024.
\newblock Chicago Crimes - 2014.
\newblock Chicago Data Portal.
\newblock Accessed: 2024-07-31.

\bibitem[{{Cincinnati Police Department}(2024)}]{cincinnaticrimeincidents2023}
{Cincinnati Police Department}. 2024.
\newblock PDI (Police Data Initiative) Crime Incidents.
\newblock City of Cincinnati Open Data Portal.
\newblock Accessed: 2024-08-01.

\bibitem[{{City of Chicago}(2024)}]{chicagotnp2023}
{City of Chicago}. 2024.
\newblock Transportation Network Providers - Trips (2023-).
\newblock Accessed: 2024-7-25.

\bibitem[{Clarke et~al.(2024)Clarke, Lim, Gupte et~al.}]{city_dangue}
Clarke, J.; Lim, A.; Gupte, P.; et~al. 2024.
\newblock A global dataset of publicly available dengue case count data.
\newblock \emph{Scientific Data}, 11: 296.

\bibitem[{Daras et~al.(2019)Daras, Green, Davies, Barr, and Singleton}]{daras_open_2019}
Daras, K.; Green, M.~A.; Davies, A.; Barr, B.; and Singleton, A. 2019.
\newblock Open data on health-related neighbourhood features in {Great} {Britain}.
\newblock \emph{Sci Data}, 6.

\bibitem[{Davison, Feldman, and Rush(2019)}]{davison2019commonsense}
Davison, J.; Feldman, J.; and Rush, A.~M. 2019.
\newblock Commonsense knowledge mining from pretrained models.
\newblock In \emph{Proceedings of the 2019 conference on empirical methods in natural language processing and the 9th international joint conference on natural language processing (EMNLP-IJCNLP)}, 1173--1178.

\bibitem[{{Department of For Hire Vehicles}(2022)}]{dctaxitrips2017}
{Department of For Hire Vehicles}. 2022.
\newblock Taxi Trips in 2017.
\newblock Accessed: 2024-7-25.

\bibitem[{Dhingra et~al.(2022)Dhingra, Cole, Eisenschlos, Gillick, Eisenstein, and Cohen}]{dhingra2022time}
Dhingra, B.; Cole, J.~R.; Eisenschlos, J.~M.; Gillick, D.; Eisenstein, J.; and Cohen, W.~W. 2022.
\newblock Time-aware language models as temporal knowledge bases.
\newblock \emph{Transactions of the Association for Computational Linguistics}, 10: 257--273.

\bibitem[{Du et~al.(2024)Du, Kang, Liu et~al.}]{city_methane}
Du, M.; Kang, X.; Liu, Q.; et~al. 2024.
\newblock City-level livestock methane emissions in China from 2010 to 2020.
\newblock \emph{Scientific Data}, 11: 251.

\bibitem[{{European Environment Agency}(2021)}]{airquality2021}
{European Environment Agency}. 2021.
\newblock Air Quality Health Risk Assessments for cities and urban centres.
\newblock Accessed: 2024-7-30.

\bibitem[{Gruver et~al.(2024)Gruver, Finzi, Qiu, and Wilson}]{gruver2024large}
Gruver, N.; Finzi, M.; Qiu, S.; and Wilson, A.~G. 2024.
\newblock Large language models are zero-shot time series forecasters.
\newblock \emph{Advances in Neural Information Processing Systems}, 36.

\bibitem[{Guo et~al.(2024)Guo, Zhang, Jiang, Peng, Yang, and Zhu}]{guo2024towards}
Guo, X.; Zhang, Q.; Jiang, J.; Peng, M.; Yang, H.~F.; and Zhu, M. 2024.
\newblock Towards Responsible and Reliable Traffic Flow Prediction with Large Language Models.
\newblock \emph{Available at SSRN 4805901}.

\bibitem[{Han et~al.(2023)Han, Xia, Xi et~al.}]{Han2023}
Han, Z.; Xia, T.; Xi, Y.; et~al. 2023.
\newblock Healthy Cities, A comprehensive dataset for environmental determinants of health in England cities.
\newblock \emph{Scientific Data}, 10.

\bibitem[{Hu et~al.(2021)Hu, Wallis, Allen-Zhu, Li, Wang, Wang, Chen et~al.}]{hulora}
Hu, E.~J.; Wallis, P.; Allen-Zhu, Z.; Li, Y.; Wang, S.; Wang, L.; Chen, W.; et~al. 2021.
\newblock LoRA: Low-Rank Adaptation of Large Language Models.
\newblock In \emph{International Conference on Learning Representations}.

\bibitem[{Jiang et~al.(2020)Jiang, Xu, Araki, and Neubig}]{jiang2020can}
Jiang, Z.; Xu, F.~F.; Araki, J.; and Neubig, G. 2020.
\newblock How can we know what language models know?
\newblock \emph{Transactions of the Association for Computational Linguistics}, 8: 423--438.

\bibitem[{Jin et~al.(2023)Jin, Wang, Ma, Chu, Zhang, Shi, Chen, Liang, Li, Pan et~al.}]{jintime}
Jin, M.; Wang, S.; Ma, L.; Chu, Z.; Zhang, J.~Y.; Shi, X.; Chen, P.-Y.; Liang, Y.; Li, Y.-F.; Pan, S.; et~al. 2023.
\newblock Time-LLM: Time Series Forecasting by Reprogramming Large Language Models.
\newblock In \emph{The Twelfth International Conference on Learning Representations}.

\bibitem[{Jin et~al.(2024)Jin, Zhang, Chen, Zhang, Liang, Yang, Wang, Pan, and Wen}]{jin2024position}
Jin, M.; Zhang, Y.; Chen, W.; Zhang, K.; Liang, Y.; Yang, B.; Wang, J.; Pan, S.; and Wen, Q. 2024.
\newblock Position: What Can Large Language Models Tell Us about Time Series Analysis.
\newblock In \emph{Forty-first International Conference on Machine Learning}.

\bibitem[{Kipf and Welling(2016)}]{Thomas2016}
Kipf, T.~N.; and Welling, M. 2016.
\newblock Semi-Supervised Classification with Graph Convolutional Networks.
\newblock \emph{CoRR}, abs/1609.02907.

\bibitem[{Li et~al.(2023)Li, Song, Liu, Ouyang, Mao, Lu, Liu, Liu, Chen, and Liu}]{li2023product}
Li, X.; Song, L.; Liu, Q.; Ouyang, X.; Mao, T.; Lu, H.; Liu, L.; Liu, X.; Chen, W.; and Liu, G. 2023.
\newblock Product, building, and infrastructure material stocks dataset for 337 Chinese cities between 1978 and 2020.
\newblock \emph{Scientific Data}, 10(1): 228.

\bibitem[{Lin et~al.(2017)Lin, Feng, dos Santos, Yu, Xiang, Zhou, and Bengio}]{Lin2017}
Lin, Z.; Feng, M.; dos Santos, C.~N.; Yu, M.; Xiang, B.; Zhou, B.; and Bengio, Y. 2017.
\newblock A Structured Self-attentive Sentence Embedding.
\newblock \emph{CoRR}, abs/1703.03130.

\bibitem[{Long et~al.(2021)Long, Jiang, Chen, Sharifi, Gasparatos, Wu, Kenichiro, guan, Yoshida, and Shigetomi}]{Long2021}
Long, Y.; Jiang, Y.; Chen, P.; Sharifi, A.; Gasparatos, A.; Wu, Y.; Kenichiro, K.; guan, D.; Yoshida, Y.; and Shigetomi, Y. 2021.
\newblock {Monthly direct and indirect greenhouse gases emissions from household consumption in the major Japanese cities}.
\newblock \emph{Scientific Data}.

\bibitem[{Longpre et~al.(2023)Longpre, Hou, Vu, Webson, Chung, Tay, Zhou, Le, Zoph, Wei et~al.}]{longpre2023flan}
Longpre, S.; Hou, L.; Vu, T.; Webson, A.; Chung, H.~W.; Tay, Y.; Zhou, D.; Le, Q.~V.; Zoph, B.; Wei, J.; et~al. 2023.
\newblock The flan collection: Designing data and methods for effective instruction tuning.
\newblock In \emph{International Conference on Machine Learning}, 22631--22648. PMLR.

\bibitem[{Manvi et~al.(2023)Manvi, Khanna, Mai, Burke, Lobell, and Ermon}]{manvi2023geollm}
Manvi, R.; Khanna, S.; Mai, G.; Burke, M.; Lobell, D.; and Ermon, S. 2023.
\newblock Geollm: Extracting geospatial knowledge from large language models.
\newblock \emph{arXiv preprint arXiv:2310.06213}.

\bibitem[{Minaee et~al.(2024)Minaee, Mikolov, Nikzad, Chenaghlu, Socher, Amatriain, and Gao}]{minaee2024large}
Minaee, S.; Mikolov, T.; Nikzad, N.; Chenaghlu, M.; Socher, R.; Amatriain, X.; and Gao, J. 2024.
\newblock Large language models: A survey.
\newblock \emph{arXiv preprint arXiv:2402.06196}.

\bibitem[{Nangini et~al.(2019)Nangini, Peregon, Ciais, Weddige, Vogel, Wang, Bréon, Bachra, Wang, Gurney, Yamagata, Appleby, Telahoun, Canadell, Grübler, Dhakal, and Creutzig}]{nangini_global_2019}
Nangini, C.; Peregon, A.; Ciais, P.; Weddige, U.; Vogel, F.; Wang, J.; Bréon, F.-M.; Bachra, S.; Wang, Y.; Gurney, K.; Yamagata, Y.; Appleby, K.; Telahoun, S.; Canadell, J.~G.; Grübler, A.; Dhakal, S.; and Creutzig, F. 2019.
\newblock A global dataset of {CO2} emissions and ancillary data related to emissions for 343 cities.
\newblock \emph{Sci Data}, 6: 180280.

\bibitem[{{New York City Taxi \& Limousine Commission}(2024)}]{nyctlctripdata}
{New York City Taxi \& Limousine Commission}. 2024.
\newblock TLC Trip Record Data.
\newblock Accessed: 2024-6-22.

\bibitem[{{NHIT Data Fusion Center}(2024)}]{DCcrimeincidents2020}
{NHIT Data Fusion Center}. 2024.
\newblock Washington D.C. Crime Incidents - 2020.
\newblock NHIT Data Fusion Center.
\newblock Accessed: 2024-07-18.

\bibitem[{nyc opendata(2023)}]{NYC311_2023}
nyc opendata. 2023.
\newblock 311 Service Requests from 2010 to Present.
\newblock \url{https://data.cityofnewyork.us/Social-Services/311-Service-Requests-from-2010-to-Present/erm2-nwe9/data_preview}.

\bibitem[{of~Austin Open Data~Portal(2024)}]{Austin311_2024}
of~Austin Open Data~Portal, C. 2024.
\newblock Austin 311 Service Requests.
\newblock \url{https://datahub.austintexas.gov/Utilities-and-City-Services/Austin-311-Public-Data/xwdj-i9he/about_data}.

\bibitem[{of~Cincinnati Open Data~Portal(2024)}]{Cincinnati311_2024}
of~Cincinnati Open Data~Portal, C. 2024.
\newblock Cincinnati 311 Non-Emergency Service Requests.
\newblock \url{https://data.cincinnati-oh.gov/Thriving-Neighborhoods/Cincinnati-311-Non-Emergency-Service-Requests/4cjh-bm8b/about_data}.

\bibitem[{Petroni et~al.(2019)Petroni, Rockt{\"a}schel, Lewis, Bakhtin, Wu, Miller, and Riedel}]{petroni2019language}
Petroni, F.; Rockt{\"a}schel, T.; Lewis, P.; Bakhtin, A.; Wu, Y.; Miller, A.~H.; and Riedel, S. 2019.
\newblock Language models as knowledge bases?
\newblock \emph{arXiv preprint arXiv:1909.01066}.

\bibitem[{{Police Department (NYPD)}(2024)}]{nypdarrestdata}
{Police Department (NYPD)}. 2024.
\newblock NYPD Arrests Data (Historic).
\newblock NYC OpenData.
\newblock Accessed: 2024-04-24.

\bibitem[{Qin and Eisner(2021)}]{qin2021learning}
Qin, G.; and Eisner, J. 2021.
\newblock Learning how to ask: Querying LMs with mixtures of soft prompts.
\newblock \emph{arXiv preprint arXiv:2104.06599}.

\bibitem[{Radford et~al.(2018)Radford, Narasimhan, Salimans, Sutskever et~al.}]{radford2018improving}
Radford, A.; Narasimhan, K.; Salimans, T.; Sutskever, I.; et~al. 2018.
\newblock Improving language understanding by generative pre-training.
\newblock \emph{OpenAI}.

\bibitem[{Roberts, Raffel, and Shazeer(2020)}]{roberts2020much}
Roberts, A.; Raffel, C.; and Shazeer, N. 2020.
\newblock How much knowledge can you pack into the parameters of a language model?
\newblock \emph{arXiv preprint arXiv:2002.08910}.

\bibitem[{Sung et~al.(2021)Sung, Lee, Yi, Jeon, Kim, and Kang}]{sung2021can}
Sung, M.; Lee, J.; Yi, S.; Jeon, M.; Kim, S.; and Kang, J. 2021.
\newblock Can language models be biomedical knowledge bases?
\newblock \emph{arXiv preprint arXiv:2109.07154}.

\bibitem[{{TomTom International BV }(2023)}]{tomtom2023}
{TomTom International BV }. 2023.
\newblock TomTom Traffic Index.
\newblock Accessed: 2024-7-31.

\bibitem[{Verbavatz and Barthelemy(2020)}]{verbavatz2020access}
Verbavatz, V.; and Barthelemy, M. 2020.
\newblock Access to mass rapid transit in OECD urban areas.
\newblock \emph{Scientific Data}, 7(1): 301.

\bibitem[{Wang et~al.(2016)Wang, Kifer, Graif, and Li}]{wang2016crime}
Wang, H.; Kifer, D.; Graif, C.; and Li, Z. 2016.
\newblock Crime rate inference with big data.
\newblock In \emph{Proceedings of the 22nd ACM SIGKDD international conference on knowledge discovery and data mining}, 635--644.

\bibitem[{Wei et~al.(2022)Wei, Wang, Schuurmans, Bosma, Xia, Chi, Le, Zhou et~al.}]{wei2022chain}
Wei, J.; Wang, X.; Schuurmans, D.; Bosma, M.; Xia, F.; Chi, E.; Le, Q.~V.; Zhou, D.; et~al. 2022.
\newblock Chain-of-thought prompting elicits reasoning in large language models.
\newblock \emph{Advances in neural information processing systems}, 35: 24824--24837.

\bibitem[{Xiao et~al.(2020)Xiao, Zhao, Shan, and Guan}]{Xiao2020}
Xiao, H.; Zhao, W.; Shan, Y.; and Guan, D. 2020.
\newblock {CO2 emission accounts of Russia’s constituent entities 2005-2019}.
\newblock \emph{Scientific Data}.

\bibitem[{Xue and Salim(2023)}]{xue2023promptcast}
Xue, H.; and Salim, F.~D. 2023.
\newblock Promptcast: A new prompt-based learning paradigm for time series forecasting.
\newblock \emph{IEEE Transactions on Knowledge and Data Engineering}.

\bibitem[{Yang et~al.(2024)Yang, Zhang, Cao, Gao, Wang, Yang, Jiang, Zha, and Shan}]{yang_comprehensive_2024}
Yang, G.; Zhang, G.; Cao, D.; Gao, X.; Wang, X.; Yang, S.; Jiang, P.; Zha, D.; and Shan, Y. 2024.
\newblock A comprehensive city-level final energy consumption dataset including renewable energy for {China}, 2005–2021.
\newblock \emph{Scientific Data}, 11: 738.

\bibitem[{Zhang et~al.(2024{\natexlab{a}})Zhang, Fu, Liang, Zhang, Yu, Cai, and Yao}]{zhang2024trafficgpt}
Zhang, S.; Fu, D.; Liang, W.; Zhang, Z.; Yu, B.; Cai, P.; and Yao, B. 2024{\natexlab{a}}.
\newblock Trafficgpt: Viewing, processing and interacting with traffic foundation models.
\newblock \emph{Transport Policy}, 150: 95--105.

\bibitem[{Zhang et~al.(2024{\natexlab{b}})Zhang, Shan, Zhao, Cai, Li, Guan et~al.}]{zhang2024water}
Zhang, Z.; Shan, Y.; Zhao, D.; Cai, B.; Li, X.; Guan, D.; et~al. 2024{\natexlab{b}}.
\newblock City-level water withdrawal and scarcity accounts of China.
\newblock figshare. Collection.
\newblock Accessed: 2024-7-30.

\bibitem[{Zhong, Friedman, and Chen(2021)}]{zhong2021factual}
Zhong, Z.; Friedman, D.; and Chen, D. 2021.
\newblock Factual Probing Is [MASK]: Learning vs. Learning to Recall.
\newblock In \emph{2021 Conference of the North American Chapter of the Association for Computational Linguistics: Human Language Technologies, NAACL-HLT 2021}, 5017--5033. Association for Computational Linguistics (ACL).

\bibitem[{Zhou et~al.(2023)Zhou, Niu, Sun, Jin et~al.}]{zhou2023one}
Zhou, T.; Niu, P.; Sun, L.; Jin, R.; et~al. 2023.
\newblock One fits all: Power general time series analysis by pretrained lm.
\newblock \emph{Advances in neural information processing systems}, 36: 43322--43355.

\bibitem[{Zillow(2024)}]{zillow2024}
Zillow, I. 2024.
\newblock Zillow Research Data.

\end{thebibliography}
\newpage
\setcounter{table}{0}
\appendix
\section{Task Descriptions}


\subsection{[1-4] NO\textsubscript{2}, O\textsubscript{3}, PM\textsubscript{10}, PM\textsubscript{2.5}}

\subsubsection{Data Description}

This study utilizes Air Quality Health Risk Assessments for cities and urban centers in 2021~\cite{airquality2021}, employing the ESTAT Urban Audit Cities (LAU) as the City Boundary Specification, focusing on cities within European countries. It aggregates and analyzes the Air Pollution Population Weighted Average statistics for four key air pollutants: NO\textsubscript{2}, O\textsubscript{3}, PM\textsubscript{10}, and PM\textsubscript{2.5}.

\subsubsection{Data Preprocess}
We first filter the rows where the ``Air Pollutant" is ``NO\textsubscript{2}". Then, we merge the ``City" column and the ``Country Or Territory" column into a new target column and select the ``Air Pollution Population Weighted Average [$\mu g/m^3$]" as the target column.
\subsubsection{Feature}
\begin{itemize}
    \item Population Density: How densely populated the area is? (0.0: very low, 10.0: very high)
    \item Traffic Volume: How heavy the traffic is? (0.0: very low, 10.0: very high)
    \item Industrial Activity: How much industrial activity is there, including factories and other industrial facilities? (0.0: very low, 10.0: very high)
    \item Sunlight Exposure: How much sunlight the area receives? (0.0: very low, 10.0: very high)
    \item Green Spaces: How extensive the green spaces are? (0.0: very poor, 10.0: very extensive)
    \item Regulatory Measures: How strict and effective the regulatory measures for controlling NO2 emission are? (0.0: very poor, 10.0: very strict)
\end{itemize}


\begin{table}[h!]
\centering
\caption{Correlation and p-value of NO$_2$ features.}
\begin{tabular}{ccc}
\toprule
Features              & Correlation & p-value                     \\ 
\midrule
population\_density  & 0.4759      & 0.0000                      \\ 
traffic\_volume      & 0.5284      & 0.0000                      \\ 
industrial\_activity & 0.3636      & 0.0000                      \\ 
sunlight\_exposure   & 0.1033      & 0.0047                      \\ 
green\_spaces        & -0.3116     & 0.0000                      \\ 
regulatory\_measures & -0.1861     & 0.0000                      \\ 
\bottomrule
\end{tabular}

\label{tab:no2_features}
\end{table}

\begin{table}[h!]
\centering
\caption{Correlation and p-value of O$_3$ features.}
\begin{tabular}{ccc}
\toprule
Features              & Correlation & p-value                     \\ 
\midrule
population\_density  & 0.1316      & 0.0003                      \\
traffic\_volume      & 0.1164      & 0.0014                      \\ 
industrial\_activity & -0.0040     & 0.9136                      \\ 
sunlight\_exposure   & 0.6217      & 0.0000                      \\ 
green\_spaces        & -0.3334     & 0.0000                      \\ 
regulatory\_measures & -0.2210     & 0.0000                      \\ 
\bottomrule
\end{tabular}

\label{tab:o3_features}
\end{table}

\begin{table}[h!]
\centering
\caption{Correlation and p-value of PM$_{10}$ features.}
\begin{tabular}{ccc}
\toprule
Features              & Correlation & p-value                     \\ 
\midrule
population\_density  & 0.1532      & 0.0000                      \\ 
traffic\_volume      & 0.1650      & 0.0000                      \\ 
industrial\_activity & 0.2149      & 0.0000                      \\ 
sunlight\_exposure   & 0.2798      & 0.0000                      \\ 
green\_spaces        & -0.4429     & 0.0000                      \\ 
regulatory\_measures & -0.4731     & 0.0000                      \\ 
\bottomrule
\end{tabular}

\label{tab:pm10_features}
\end{table}

\begin{table}[h!]
\centering
\caption{Correlation and p-value of PM$_{2.5}$ features.}
\begin{tabular}{ccc}
\toprule
Features              & Correlation & p-value                     \\ 
\midrule
population\_density  & 0.0415      & 0.2561                      \\ 
traffic\_volume      & 0.1015      & 0.0054                      \\ 
industrial\_activity & 0.2222      & 0.0000                      \\ 
sunlight\_exposure   & -0.0114     & 0.7555                      \\ 
green\_spaces        & -0.2834     & 0.0000                      \\ 
regulatory\_measures & -0.3841     & 0.0000                      \\ 
\bottomrule
\end{tabular}

\label{tab:pm25_features}
\end{table}

\subsection{[5] Methane}

\subsubsection{Data Description}

This dataset~\cite{city_methane} presents a comprehensive city-level inventory of livestock methane emission in China from 2010 to 2020, incorporating biological, management, and environmental variables. The data is measured in tonnes.

\subsubsection{Data Preprocess}

The methane emission from livestock, measured in tons per year, are analyzed using data from 2020, encompassing 347 cities across China.

\subsubsection{Feature}

\begin{itemize}
    \item Livestock Density: Measures the density of livestock in the area (0.0: very low density, 10.0: very high density).  Based on the number of livestock per square mile.
    \item Feeding Practices: Reflects the type of feeding practices in the area (0.0: very poor, 10.0: very good).  Based on the composition of livestock feed.
    \item Manure Management: Measures the effectiveness of manure management practices in the area (0.0: very poor, 10.0: excellent).  Includes methods of handling and processing livestock manure.
    \item Climate Conditions: Reflects the climate conditions of the area  (0.0: very low, 10.0: very high).  Based on temperature and humidity.
    \item Policy and Technology: Measures the implementation of policies and technologies aimed at reducing methane emission (0.0: very poor, 10.0: excellent).  Includes the use of methane inhibitors, feed additives, and government regulations.
\end{itemize}


\begin{table}[h!]
\centering
\caption{Correlation and p-value of Methane features.}
\begin{tabular}{ccc}
\toprule
Features & Correlation & p-value \\ 
\midrule
livestock\_density                 & 0.2377     & 0.0000    \\ 
feeding\_practices                       & -0.0517     & 0.3368   \\ 
manure\_management                     & -0.0383     & 0.4765    \\ 
climate\_conditions            & -0.0896     & 0.0956    \\ 
policy\_and\_technology                 & -0.3129     & 0.0000    \\ 
\bottomrule
\end{tabular}

\label{tab:methane_features}
\end{table}

\subsection{[6] Carbon emission}

\subsubsection{Data Description}

This dataset~\cite{nangini_global_2019} provides a comprehensive compilation of anthropogenic CO\textsubscript{2} emission data for 343 cities worldwide. It integrates data from three major sources: the Carbon Disclosure Project (CDP), the Bonn Center for Local Climate Action and Reporting (Carbonn), and Peking University. The dataset includes both self-reported and systematically collected data, with additional quality control measures and corrections applied to ensure accuracy and consistency. The data is measured in million tonnes of CO\textsubscript{2} (MtCO\textsubscript{2}-eq).

\subsubsection{Data Preprocess}
Total emission (CDP) [MtCO\textsubscript{2}-eq] is chosen as the target value. Then Cities with zero Carbon emission values are removed from thes dataset. 

\subsubsection{Feature}
\begin{itemize}
    \item Population: The population of the city. (0.0: very low, 10.0: very high)
    \item GDP: The Gross Domestic Product of the city. (0.0: very low, 10.0: very high)
    \item Average Annual Temperature: The average annual temperature of the city. (0.0: very low, 10.0: very high)
    \item Built-up Area: The built-up area of the city. (0.0: very low, 10.0: very high)
\end{itemize}

\begin{table}[h!]
\centering
\caption{Correlation and p-value of Carbon emission features.}
\begin{tabular}{ccc}
\toprule
Features                    & Correlation   & p-value        \\ 
\midrule
population                  & 0.5605        & 0.0000         \\ 
gdp                         & 0.4387        & 0.0000         \\ 
average\_annual\_temperature  & 0.1725        & 0.0186         \\ 
built-up\_area               & 0.5119        & 0.0000         \\ 
\bottomrule
\end{tabular}
\label{tab:co2_features}
\end{table}

\subsection{[7] Household CO\textsubscript{2} }

\subsubsection{Data Description}

The dataset utilized in this study~\cite{Long2021} comprises an emission inventory of urban household CO\textsubscript{2}-equivalent emissions for 52 major cities in Japan. It includes fossil fuels related to direct household emissions such as gasoline, kerosene, liquefied petroleum gas, and city gas. These emissions are calculated by converting the monthly household expenditures on these fuels, as reported in the \cite{FIES}, into grams of CO\textsubscript{2}-equivalent per capita (g-CO\textsubscript{2}eq/capita).

\subsubsection{Data Preprocess}

For the purpose of this study, the direct emission [g-CO\textsubscript{2} equivalent per capita] for the year 2015 across the 52 Japanese cities are selected as the target variable.

\subsubsection{Feature}
\begin{itemize}
    \item Use of Fossil Fuels: Household use of gasoline, kerosene, LPG, and city gas. (0.0: very low, 10.0: very high)
    \item Consumption of Goods and Services: consumed goods and services, including food, electricity, durable goods, and entertainment. (0.0: very low, 10.0: very high)
    \item City Size and Population Density: The size and population density of the city. (0.0: very low, 10.0: very high)
    \item Population: The population of the city. (0.0: very low, 10.0: very high)
    \item Household Economic Level: The economic level of the households in the city. (0.0: very low, 10.0: very high)
    \item Average Annual Temperature: The average annual temperature of the city. (0.0: very low, 10.0: very high)
    \item Public Transport and Infrastructure: The development of public transport systems and infrastructure. (0.0: very low, 10.0: very high)
\end{itemize}


\begin{table}[h!]
\centering
\caption{Correlation and p-value of Household CO\textsubscript{2} features.}
\begin{tabular}{ccc}
\toprule
Features              & Correlation & p-value        \\ 
\midrule
use\_of\_fossil\_fuels          & -0.2566     & 0.0663         \\ 
consumption & -0.2688 & 0.0540         \\ 
city\_size & -0.2262  & 0.1069         \\ 
population                   & -0.1786     & 0.2052         \\ 
household\_economy      & -0.1455    & 0.3033         \\ 
average\_annual\_temperatures  & -0.7210     & 0.0000         \\ 
public\_transport\_and\_infrastructure & -0.2193 & 0.1183     \\ 
\bottomrule
\end{tabular}
\label{tab:Japan_co2_features}
\end{table}

\subsection{[8] Total CO\textsubscript{2} }

\subsubsection{Data Description}

The dataset \cite{Xiao2020} provides CO\textsubscript{2} emissions data for Russia’s 82 constituent entities from 2005 to 2019. For this experiment, the total CO\textsubscript{2} emissions of each entity in 2019 are selected, with emissions measured in million tonnes.

\subsubsection{Data Preprocess}

The target variable is calculated by summing the energy-related and process-related emissions for each entity in 2019.

\subsubsection{Feature}
\begin{itemize}
    \item Agriculture Forestry and Fishing: How significant are agriculture, forestry, and fishing activities in the region? (0.0: very low, 10.0: very high)
    \item Mining and Quarrying: How significant are mining and quarrying activities in the region? (0.0: very low, 10.0: very high)
    \item Manufacturing: How significant are manufacturing activities in the region? (0.0: very low, 10.0: very high)
    \item Electricity, Gas, Steam: How significant are these utilities in the region? (0.0: very low, 10.0: very high)
    \item Water Supply: How significant are these activities in the region? (0.0: very low, 10.0: very high)
    \item Construction: How significant are construction activities in the region? (0.0: very low, 10.0: very high)
    \item Wholesale and Retail Trade: How significant are wholesale and retail trade activities in the region? (0.0: very low, 10.0: very high)
    \item Transportation and Storage: How significant are transportation and storage activities in the region? (0.0: very low, 10.0: very high)
    \item Human Health and Social Work: How significant are human health and social work activities in the region? (0.0: very low, 10.0: very high)    
\end{itemize}


\begin{table}[h!]
\centering
\caption{Correlation and p-value of Total CO\textsubscript{2} features.}
\begin{tabular}{ccc}
\toprule
Features              & Correlation & p-value        \\ 
\midrule
agriculture\_forestry\_fishing  & -0.2925     & 0.0077        \\ 
mining\_and\_quarrying         & 0.3091     & 0.0047         \\ 
manufacturing                & 0.4546     & 0.0000         \\ 
electricity\_gas\_steam      & 0.3130     & 0.0042         \\ 
water\_supply                 & 0.3830     & 0.0004         \\ 
construction                 & 0.3451     & 0.0015         \\ 
wholesale\_and\_retail\_trade   & 0.3082     & 0.0048         \\ 
transportation\_and\_storage   & 0.4165     & 0.0001         \\ 
human\_health\_and\_social\_work & 0.1907     & 0.0862         \\ 
\bottomrule
\end{tabular}
\label{tab:Russia_co2_features}
\end{table}

\subsection{[9] Water withdrawal} 

\subsubsection{Data Description}

The City-level water withdrawal and scarcity accounts of China dataset~\cite{zhang2024water} is used for this experiment. The data is measured in 100 million cubic meters (100 million $m^3$).

\subsubsection{Data Preprocess}
The dataset provides information on water withdrawal, availability, and criticality for different cities in China. For this task, only the water withdrawal data is considered. The zero-value entries are removed to ensure the dataset accurately reflects cities with active water usage.
\subsubsection{Feature}
\begin{itemize}
    \item Population: How densely populated the area is? (0.0: very low, 10.0: very high)
    \item Climate: The general climate of the area. (0.0: very harsh, 10.0: very mild)
    \item Economy: The economic activity and development level. (0.0: very poor, 10.0: very strong)
    \item Agriculture: The extent of agricultural activity and land use. (0.0: very low, 10.0: very high)
    \item Industry: How much industrial activity is there, including factories and other industrial facilities? (0.0: very low, 10.0: very high)
\end{itemize}

\begin{table}[h!]
\centering
\caption{Correlation and p-value of Water withdrawal features.}
\begin{tabular}{ccc}
\toprule
Features              & Correlation & p-value        \\ 
\midrule
population            & 0.3919      & 0.0000         \\ 
climate               & 0.0162      & 0.7488         \\ 
economy               & 0.3374      & 0.0000         \\ 
agriculture           & 0.1117      & 0.0264         \\ 
industry              & 0.2700      & 0.0000         \\ 
\bottomrule
\end{tabular}
\end{table}

\subsection{[10-12] Energy (Total, Traditional and Renewable)}

\subsubsection{Data Description}

The dataset~\cite{yang_comprehensive_2024} provides an extensive inventory of final energy consumption across 331 Chinese cities from 2005 to 2021. The data is measured in 10,000 tonnes of coal equivalent (10Kt-coal-eq).

\subsubsection{Data Preprocess}
Cities with zero energy consumption values are removed from the dataset. The Total, Traditional, and Renewable Energy consumption data are each selected as target variables for separate analyses.

\subsubsection{Feature}
\begin{itemize}
    \item Population: The population of the city. (0.0: very low, 10.0: very high)
    \item GDP: The Gross Domestic Product of the city. (0.0: very low, 10.0: very high)
    \item Average Annual Temperature: The average annual temperature of the city. (0.0: very low, 10.0: very high)
    \item Built-up Area: The built-up area of the city. (0.0: very low, 10.0: very high)
\end{itemize}

\begin{table}[h!]
\centering
\caption{Correlation and p-value of Total energy features.}
\begin{tabular}{ccc}
\toprule
Features                    & Correlation   & p-value        \\ 
\midrule
population                  & 0.6026        & 0.0000         \\ 
gdp                         & 0.6467        & 0.0000         \\ 
average\_annual\_temperature  & -0.0573       & 0.2983         \\ 
built-up\_area               & 0.5679        & 0.0000         \\ 
\bottomrule
\end{tabular}
\end{table}

\begin{table}[h!]
\centering
\caption{Correlation and p-value of Traditional energy features.}
\begin{tabular}{ccc}
\toprule
Features                    & Correlation   & p-value        \\ 
\midrule
population                  & 0.5938        & 0.0000         \\ 
gdp                         & 0.6026        & 0.0000         \\ 
average\_annual\_temperature  & -0.0617       & 0.1762         \\ 
built-up\_area               & 0.5464        & 0.0000         \\ 
\bottomrule
\end{tabular}
\end{table}

\begin{table}[h!]
\centering
\caption{Correlation and p-value of Renewable energy features.}
\begin{tabular}{ccc}
\toprule
Features                    & Correlation   & p-value        \\ 
\midrule
population                  & 0.6315        & 0.0000         \\ 
gdp                         & 0.6058        & 0.0000         \\ 
average\_annual\_temperature  & 0.2058        & 0.0000         \\ 
built-up\_area               & 0.5914        & 0.0000         \\ 
\bottomrule
\end{tabular}
\end{table}

\subsection{[13] Dengue}

\subsubsection{Data Description}

This paper introduces the OpenDengue dataset\cite{city_dangue}, a global collection of publicly available dengue case counts from 1924 to 2023, covering 102 countries.  The dataset integrates data from 843 sources, emphasizing its high temporal (weekly/monthly) and spatial resolution.  This resource aims to support research on dengue dynamics, enhance disease surveillance, and inform public health interventions. The data is measured in counts.

\subsubsection{Data Preprocess}
The total number of dengue fever cases in 2020 across 1,541 cities worldwide is chosen as the target variable. Cities with missing data are excluded from the analysis to ensure data completeness and consistency.

\subsubsection{Feature}
\begin{itemize}
    \item Population Density: Measures the population density of the area (0.0: very low, 10.0: very high). Based on the number of residents per square mile.
    \item Temperature: Reflects the average temperature of the area (0.0: very low, 10.0: very high). Based on the annual average temperature.
    \item Rainfall: Reflects the annual rainfall level in the area (0.0: very low, 10.0: very high). Based on the annual average rainfall.
    \item Public Health Measures: Measures the effectiveness of public health measures in the area (0.0: very low, 10.0: very high). Includes mosquito control, public education, and healthcare infrastructure.
    \item Urbanization Level: Measures the level of urbanization in the area (0.0: very low, 10.0: very high). Based on the presence of urban infrastructure, services, and population concentration.
\end{itemize}

\begin{table}[h!]
\centering
\caption{Correlation and p-value of Dengue features.}
\begin{tabular}{ccc}
\toprule
Features    & Correlation & p-value        \\ 
\midrule
population\_density         & 0.2867      & 0.0000         \\ 
temperature                 & 0.0259      & 0.3105         \\ 
rainfall                    & 0.0701      & 0.0059         \\ 
public\_health\_measures     & 0.1451      & 0.0000         \\ 
urbanization\_level         & 0.2908      & 0.0000         \\ 
\bottomrule
\end{tabular}
\label{tab:dengue_features}
\end{table}

\subsection{[14] COVID-19}
\subsubsection{Data Description}

This dataset~\cite{cdc_covid_death_counts} presents a provisional count of deaths involving COVID-19 by county of occurrence in the United States from 2020 to 2023. The data is measured in counts.

\subsubsection{Data Preprocess}
Among the six urban-rural codes (Noncore, Medium metro, Small metro, Micropolitan, Large fringe metro, Large central metro), only Large fringe metro and large central metro are reserved and zero values are dropped.

\subsubsection{Feature}

\begin{itemize}
    \item Population Density: How densely populated the county is? (0.0: very low, 10.0: very high)
    \item Age Distribution: What percentage of the county's population is elderly (65+ years old)? (0.0: very low, 10.0: very high)
    \item Healthcare Capacity: How well-equipped is the county's healthcare system? (0.0: very poor, 10.0: excellent)
    \item Prevalence of Pre-existing Conditions: What percentage of the population has chronic illnesses? (0.0: very low, 10.0: very high)
    \item Socioeconomic Factors: How favorable are the socioeconomic conditions in the county? (0.0: very unfavorable, 10.0: very favorable)
\end{itemize}

\begin{table}[h!]
\centering
\caption{Correlation and p-value of COVID-19 features.}
\begin{tabular}{ccc}
\toprule
Features             & Correlation & p-value        \\ 
\midrule
population          & 0.4238      & 0.0000         \\ 
age                 & -0.0728     & 0.0679         \\ 
healthcare          & 0.2558      & 0.0000         \\ 
existing            & 0.1927      & 0.0000         \\ 
social              & 0.0999      & 0.0122         \\ 
\bottomrule
\end{tabular}
\label{tab:covid_features}
\end{table}

\subsection{[15] Life expectancy}

\subsubsection{Data Description}

This dataset~\cite{Han2023} is used to study the impact of environmental factors on health in English cities. It includes a variety of variables related to the environment and health, which can be used to analyze the relationship between health conditions in different cities and environmental factors. The data is measured in years.

\subsubsection{Data Preprocess}

We have chosen healthy life expectancy at birth as the primary target variable for our research, as it provides a comprehensive measure of population health. To ensure the accuracy and reliability of our analysis, we meticulously cleaned the dataset by removing all instances of missing or NaN values.

\subsubsection{Feature}

\begin{itemize}
    \item Healthcare Quality: Reflects the overall quality and accessibility of healthcare services in the area (0.0: very poor, 10.0: excellent).
    \item Economic Status: Measures the economic development and average income levels of residents (0.0: very low, 10.0: very high).
    \item Public Health Infrastructure: Reflects the availability and quality of public health facilities and services (0.0: very poor, 10.0: excellent).
    \item Education Level: Measures the average educational attainment and quality of education services in the area (0.0: very low, 10.0: very high).
    \item Environmental Conditions: Reflects the overall environmental quality, including air and water quality (0.0: very poor, 10.0: excellent).
\end{itemize} 

\begin{table}[h!]
\centering
\caption{Correlation and p-value of Life expectancy features.}
\begin{tabular}{ccc}
\toprule
Features                        & Correlation & p-value        \\ 
\midrule
healthcare\_quality            & 0.6480      & 0.0001         \\ 
economic\_status               & 0.6381      & 0.0002         \\ 
public\_health\_infrastructure & 0.5385      & 0.0026         \\ 
education\_level               & 0.7215      & 0.0000         \\ 
environmental\_conditions      & 0.6200      & 0.0003         \\ 
\bottomrule
\end{tabular}
\label{tab:life_expectancy}
\end{table}

\subsection{[16] Health access}

\subsubsection{Data Description}
This dataset~\cite{daras_open_2019} provides a comprehensive series of national open source low-level geographical measures of accessibility to various health-related features across Great Britain. It includes 14 measures categorized into three domains: retail environment, health services, and physical environment. The dataset is created using network analysis to calculate postcode accessibility to these features, resulting in a valuable resource for researchers and policymakers interested in the spatial determinants of health.

\subsubsection{Data Preprocess}
We select ``Access to Healthy Assets and Hazards'' (AHAH) as our target variable. Outliers and missing values are removed from the dataset.

\subsubsection{Feature}
\begin{itemize}
    \item Retail Environment: The availability and type of retail establishments, such as fast food outlets and pubs, can significantly influence the dietary and social health of a community. (0.0: very low, 10.0: very high)
    \item Physical Environment: The presence of green spaces, pollution levels, and other physical factors can affect the overall health and well-being of residents. (0.0: very low, 10.0: very high)
    \item population: The population of the city. (0.0: very low, 10.0: very high)
    \item GDP: The Gross Domestic Product of the city. (0.0: very low, 10.0: very high)
    \item average annual temperature: The average annual temperature of the city. (0.0: very low, 10.0: very high)
    \item built-up area: The built-up area of the city. (0.0: very low, 10.0: very high)
\end{itemize}

\begin{table}[h!]
\centering
\caption{Correlation and p-value of Health access features.}
\begin{tabular}{ccc}
\toprule
Features                    & Correlation   & p-value        \\ 
\midrule
retail\_environment          & 0.3120        & 0.0000         \\ 
physical\_environment        & -0.3968       & 0.0000         \\ 
population                  & 0.4070        & 0.0000         \\ 
gdp                         & 0.2713        & 0.0000         \\ 
average\_annual\_temperature  & 0.0789        & 0.1278         \\ 
built-up\_area               & 0.4439        & 0.0000         \\ 
\bottomrule
\end{tabular}
\end{table}

\subsection{[17-18] Avg travel time, Avg speed}

\subsubsection{Data Description}

The TomTom Traffic Index website~\cite{tomtom2023} provides rankings and analysis of traffic information in 387 cities worldwide. Key data metrics include:
\begin{itemize}
    \item Average travel time per 10 km: The average time to travel 10 kilometers, shown in minutes and seconds. (0.0: very low, 10.0: very high)
    \item Change from 2022: The difference in travel time compared to 2022. (0.0: very low, 10.0: very high)
    \item Congestion level: The percentage of extra travel time due to congestion during peak hours. (0.0: very low, 10.0: very high)
    \item Time lost per year at rush hours: The total hours lost annually in rush hour traffic. (0.0: very low, 10.0: very high)
    \item Average speed in rush hour: The average speed during peak hours, in kilometers per hour. (0.0: very low, 10.0: very high)
\end{itemize}
This study selects the average travel time per 10 km and average speed in rush hour for analysis. The data is measured in seconds.

\subsubsection{Data Preprocess}
We convert the data in the ``Average travel time per 10 km" column, such as ``37 min 20s," into seconds for the target column. Additionally, we remove the ``km/h" unit from the ``Average speed in rush hour" column and use the modified values as the target column. The ``City" column is used as the zone column.

\subsubsection{Feature}  
\begin{itemize}
    \item Traffic Volume: How heavy the traffic is? (0.0: very low, 10.0: very high)
    \item Road Network Quality: How good the road infrastructure is, including the number and condition of roads, highways, and bridges? (0.0: very poor, 10.0: excellent)
    \item  Traffic Signal Efficiency: How well-optimized and efficient the traffic signal system is, including signal timing and coordination? (0.0: very poor, 10.0: very efficient)
    \item Public Transportation: How extensive and efficient the public transportation system is, including buses, trains, and subways? (0.0: very poor, 10.0: very extensive and efficient)
    \item Urban Planning: How well the city is planned, including the distribution of residential, commercial, and industrial areas, as well as zoning and land use? (0.0: very poor, 10.0: very well planned)
\end{itemize}

\begin{table}[h!]
\centering
\caption{Correlation and p-value of Avg travel time features.}
\begin{tabular}{ccc}
\toprule
Features & Correlation & p-value        \\ 
\midrule
traffic\_volume         & 0.3224      & 0.0000         \\ 
road\_network\_quality  & -0.0200     & 0.6953         \\ 
traffic\_signal\_efficiency & 0.0483  & 0.3436         \\ 
public\_transportation  & 0.5529      & 0.0000         \\ 
urban\_planning         & 0.1008      & 0.0476         \\ 
\bottomrule
\end{tabular}
\label{tab:avg_travel_time_features}
\end{table}

\begin{table}[h!]
\centering
\caption{Correlation and p-value of Avg speed features.}
\begin{tabular}{ccc}
\toprule
Features & Correlation & p-value        \\ 
\midrule
traffic\_volume         & -0.2805     & 0.0000         \\ 
road\_network\_quality  & -0.0339     & 0.5063         \\ 
traffic\_signal\_efficiency & -0.0956  & 0.0601         \\ 
public\_transportation  & -0.6363     & 0.0000         \\ 
urban\_planning         & -0.1725     & 0.0007         \\ 
\bottomrule
\end{tabular}
\label{tab:avg_speed_rush_hour_features}
\end{table}

\subsection{[19-21] PNT (500m, 1000m and 1500m)}

\subsubsection{Data Description}

The People Near Transit (PNT) levels are defined as the share of urban population living at geometric distances of 500m, 1,000m, and 1,500m from any MRT (mass rapid transit) station in the agglomeration. This paper~\cite{verbavatz2020access} introduces the largest global dataset of PNT. In their definition, mass rapid transit encompasses:
\begin{itemize}
    \item Tram, streetcar, or light rail services.
    \item Subway, Metro, or any underground service.
    \item Suburban rail services.
\end{itemize}
The data is measured in percentage (\%).

\subsubsection{Data Preprocess}
We merge the ``City" column and the ``Country" column into a new target column, and select the ``500m", ``1000m" and ``1500m" as the target columns.

\subsubsection{Feature}
\begin{itemize}
    \item Urban Planning: How well the city is planned, including the distribution of residential, commercial, and industrial areas, as well as zoning and land use? (0.0: very poor, 10.0: very well planned)
    \item Road Network Quality: How good the road infrastructure is, including the number and condition of roads, highways, and bridges? (0.0: very poor, 10.0: excellent)
    \item Public Transportation: How extensive and efficient the public transportation system is, including buses, trains, and subways? (0.0: very poor, 10.0: very extensive and efficient)
    \item Government Policy: How is the government's investment and development strategy for public transportation? (0.0: very poor, 10.0: very good)
\end{itemize}

\begin{table}[h!]
\centering
\caption{Correlation and p-value of PNT(500m) features.}
\begin{tabular}{ccc}
\toprule
Features  & Correlation & p-value        \\ 
\midrule
urban\_planning             & 0.5208      & 0.0000         \\ 
road\_network\_quality      & 0.3495      & 0.0010         \\ 
public\_transportation      & 0.6791      & 0.0000         \\ 
government\_policy          & 0.5000      & 0.0000         \\ 
\bottomrule
\end{tabular}
\label{tab:pnt_level_500m_features}
\end{table}

\begin{table}[h!]
\centering
\caption{Correlation and p-value of PNT(1000m) features.}
\begin{tabular}{ccc}
\toprule
Features & Correlation & p-value        \\ 
\midrule
urban\_planning             & 0.6005      & 0.0000         \\ 
road\_network\_quality      & 0.4185      & 0.0001         \\ 
public\_transportation      & 0.7611      & 0.0000         \\ 
government\_policy          & 0.5862      & 0.0000         \\ 
\bottomrule
\end{tabular}
\label{tab:pnt_level_1000m_features}
\end{table}

\begin{table}[h!]
\centering
\caption{Correlation and p-value of PNT(1500m) features.}
\begin{tabular}{ccc}
\toprule
Features & Correlation & p-value        \\ 
\midrule
urban\_planning             & 0.6327      & 0.0000         \\ 
road\_network\_quality      & 0.4408      & 0.0000         \\ 
public\_transportation      & 0.7895      & 0.0000         \\ 
government\_policy          & 0.6192      & 0.0000         \\ 
\bottomrule
\end{tabular}
\label{tab:pnt_level_1500m_features}
\end{table}

\subsection{[22-25] Industry (Total, Mining, Manufacture, Utilities)} 
\subsubsection{Data Description}

This dataset~\cite{zhang2024water} contains data including various types of industry output from 245 cities in China in 2015. The data is measured in 100 million Yuan.

\subsubsection{Data Preprocess}
The industry categorization is divided into three sectors: Mining (e.g., coal, metal ores, petroleum extraction), Manufacture (including food processing, textiles, machinery, electronics, and more specialized manufacturing), and Utilities (comprising electricity, gas, and water supply services).

    
    

\subsubsection{Feature}
\begin{itemize}
    \item Population: How densely populated the area is? (0.0: very low, 10.0: very high)
    \item Resource: The amount of resources of the area (0.0: very scarce, 10.0: very rich)
    \item Economy: The economic activity and development level (0.0: very poor, 10.0: very strong)
    \item Infrastructure: The level of infrastructure, including factories and industrial facilities (0.0: very low, 10.0: very high)
    \item Technology: The level of technology (0.0: very low, 10.0: very high)
\end{itemize}
    
\begin{table}[h!]
\centering
\caption{Correlation and p-value of Total industry features.}
\begin{tabular}{ccc}
\toprule
Features         & Correlation & p-value \\
\midrule
population      & 0.6394      & 0.0000  \\
resource        & -0.1225     & 0.0556  \\
economy         & 0.7632      & 0.0000  \\
infrastructure  & 0.6840      & 0.0000  \\
technology      & 0.7245      & 0.0000  \\
\bottomrule
\end{tabular}
\label{tab:Industry_feature_correlations}
\end{table}

\begin{table}[h!]
\centering
\caption{Correlation and p-value of Mining features.}
\begin{tabular}{ccc}
\toprule
Features         & Correlation & p-value \\
\midrule
population      & -0.0031     & 0.9611  \\
resource        & 0.3724      & 0.0000  \\
economy         & 0.1105      & 0.0843  \\
infrastructure  & 0.1096      & 0.0869  \\
technology      & 0.0821      & 0.2004  \\
\bottomrule
\end{tabular}
\label{tab:Mining_feature_correlations}
\end{table}

\begin{table}[h!]
\centering
\caption{Correlation and p-value of Manufacture features.}
\begin{tabular}{ccc}
\toprule
Features         & Correlation & p-value \\
\midrule
population      & 0.6361      & 0.0000  \\
resource        & -0.1448     & 0.0234  \\
economy         & 0.7505      & 0.0000  \\
infrastructure  & 0.6750      & 0.0000  \\
technology      & 0.7138      & 0.0000  \\
\bottomrule
\end{tabular}
\label{tab:Manufacture_feature_correlations}
\end{table}

\begin{table}[h!]
\centering
\caption{Correlation and p-value of Utilities features.}
\begin{tabular}{ccc}
\toprule
Features         & Correlation & p-value \\
\midrule
population      & 0.5008      & 0.0000  \\
resource        & -0.1034     & 0.1063  \\
economy         & 0.6244      & 0.0000  \\
infrastructure  & 0.5233      & 0.0000  \\
technology      & 0.5925      & 0.0000  \\
\bottomrule
\end{tabular}
\label{tab:Utilities_feature_correlations}
\end{table}

\subsection{[26] Patent} 
\subsubsection{Data Description}

This dataset~\cite{deRassenfosse2019patent} provides geographic coordinates for inventor and applicant locations in 18.8 million patent documents spanning more than 30 years. The geocoded data is further allocated to the corresponding countries, regions, and cities. The data is measured in counts.

\subsubsection{Data Preprocess}
Data from 2014 is selected, focusing on the top 500 cities. Due to errors in the names of 13 cities, the final dataset includes 487 cities.

\subsubsection{Feature}
\begin{itemize}
    \item     Population: How densely populated the city is? (0.0: very low, 10.0: very high)
	\item Education: The amount of educational resource in this city. (0.0: very scarce, 10.0: very rich)
	\item Economy: The economic activity and development level. (0.0: very poor, 10.0: very strong)
	\item Policy: Government Incentives and Intellectual Property Laws. (0.0: very low, 10.0: very high)
	\item Technology: The level of technology. (0.0: very low, 10.0: very high)
\end{itemize}

\begin{table}[h!]
\centering
\caption{Correlation and p-value of Patent features.}
\begin{tabular}{ccc}
\toprule
Features         & Correlation & p-value \\
\midrule
population      & 0.0645      & 0.1556  \\
education       & 0.1497      & 0.0009  \\
economy         & 0.1814      & 0.0001  \\
policy          & 0.1235      & 0.0064  \\
technology      & 0.1875      & 0.0000  \\
\bottomrule
\end{tabular}
\label{tab:Patent Features}
\end{table}

\subsection{[27] Avg home value}

\subsubsection{Data Description}

This dataset~\cite{zillow2024} includes housing data with home value information derived from more than 374 million public records and assessor data for approximately 200 million parcels across over 3,100 counties in the United States. The values are measured in thousands of U.S. dollars.

\subsubsection{Data Preprocess}

The top 200 largest U.S. cities, ranked by size in the dataset, are selected for this study. The average monthly urban housing prices for each city in 2023 are averaged to represent the annual average housing price, which is used as the target variable.

\subsubsection{Feature}
\begin{itemize}
    \item Regional Economic Strength: How strong is the overall economy of the city, including major industries and employment opportunities? (0.0: very low, 10.0: very high).
    \item Population Growth and Density: How significant is the population growth and density in the city? (0.0: very low, 10.0: very high).
    \item Cost of living and quality of life: How does the cost of living and quality of life (including climate, healthcare, and education) in the city impact home values? (0.0: very low, 10.0: very high).
    \item Infrastructure and public services: How well-developed are the infrastructure and public services (transportation, healthcare, education) in the city? (0.0: very low, 10.0: very high).
    \item local\_housing\_market\_dynamics: How strong are the local housing market trends and housing supply dynamics in the city? (0.0: very low, 10.0: very high).
\end{itemize}


\begin{table}[h!]
\centering
\caption{Correlation and p-value of Avg home value features.}
\begin{tabular}{ccc}
\toprule
Features                                & Correlation & p-value        \\ 
\midrule
regional\_economic\_strength            & 0.4283      & 0.0000         \\ 
population\_growth\_and\_density        & 0.4689      & 0.0000         \\ 
cost\_of\_living\_and\_quality\_of\_life & 0.3987      & 0.0000         \\ 
infrastructure\_and\_public\_services   & 0.3792      & 0.0000         \\ 
local\_housing\_market\_dynamics        & 0.5770      & 0.0000         \\ 
\bottomrule
\end{tabular}
\label{tab:zillow_home_value_feature}
\end{table}

\subsection{[28] Material stocks}

\subsubsection{Data Description}

This paper~\cite{li2023product} introduces a dataset that contains the stock of 24 materials contained in 10 types of products, buildings, and infrastructure in all 332 prefecture-level cities in China from 1978 to 2020. We select the per capita material stocks as our research target for this part. The data is measured in tonnes per capita (tonnes/capita).

\subsubsection{Data Preprocess}
We select the data from the ``2020" column in the ``Material stock and population 1978-2020" table, add ``China" to the ``City" column to create the zone column and use the ``2020" column as the target column. 

\subsubsection{Feature}
\begin{itemize}
    \item Economic Activity: How high is the level of economic activity, including industrial output, commercial activity, and service sector strength? (0.0: very low, 10.0: very high)
    \item Infrastructure Quality: How good is the city's infrastructure, including transportation, utilities, and communication systems? (0.0: very poor, 10.0: excellent)
    \item Housing Development: How well-developed is the housing sector, including the availability and quality of residential buildings? (0.0: very poor, 10.0: very well-developed)
    \item Land Use Policy: How effective are the city's land use policies, including zoning regulations, green space allocation, and land management practices? (0.0: very poor, 10.0: very effective)
\end{itemize}

\begin{table}[h!]
\centering
\caption{Correlation and p-value of Material stocks features.}
\begin{tabular}{ccc}
\toprule
Features  & Correlation & p-value        \\ 
\midrule
economic\_activity         & -0.3082     & 0.0000         \\ 
infrastructure\_quality    & -0.2871     & 0.0000         \\ 
housing\_development       & -0.2862     & 0.0000         \\ 
land\_use\_policy          & -0.1278     & 0.0198         \\ 
\bottomrule
\end{tabular}
\label{tab:per_capita_material_stocks_features}
\end{table}

\subsection{[29-34] Transport}
\subsubsection{Data Description}

For this experiment, we use trip record data from New York City, Chicago, and Washington D.C., normalizing the data across all cities to a count scale ranging from 0 to 10. The New York City data comes from the Taxi and Limousine Commission (TLC) trip record data collected in 2024~\cite{nyctlctripdata}. For the Chicago analysis, we utilize the Transportation Network Providers (TNP) Trips dataset, which is provided by the City of Chicago and last updated on July 26, 2024~\cite{chicagotnp2023}. In Washington D.C., the experiment utilizes data from the Taxi Trips in 2017 dataset, which is provided by the Department of For-Hire Vehicles and last updated on September 27, 2022~\cite{dctaxitrips2017}.

\subsubsection{Data Preprocess}

The study processes for-hire vehicle data from New York City, Chicago, and Washington D.C., focusing on hourly weekday trip patterns. The data is filtered to retain only weekday trips and is grouped by hour, pick-up zone, and drop-off zone to calculate counts, which are normalized on a scale of 0.0 to 10.0. The processed dataset includes hourly pick-up and drop-off counts, normalized within each zone, with missing values filled in as zeros. The final dataset provides a comprehensive view of hourly traffic flow in each zone for weekdays, facilitating further analysis.

\subsubsection{Feature}

\begin{itemize}
    \item Residential: How residential the area is? (0.0: least residential zone, 10.0: most residential zone)
\item Commercial: How commercial the area is? (0.0: least commercial zone, 10.0: most commercial zone)
\item Recreation: How people would come here for recreation? (0.0: least recreational zone, 10.0: most recreational zone)
\item Tourism: How much tourism is there? (0.0: least tourists zone, 10.0: most tourists zone)
\item Transportation: How good is the transportation in the area? (0.0: worst transportation zone, 10.0: best transportation zone)
\end{itemize}

\begin{table}[h!]
\centering
\caption{Correlation and p-value of NYC pick-up features.}
\begin{tabular}{ccc}
\toprule
Features         & Correlation & p-value \\
\midrule
hour            & 0.5003      & 0.0000  \\
residential     & 0.0245      & 0.0544  \\
commercial      & 0.0442      & 0.0005  \\
recreation      & 0.0074      & 0.5609  \\
tourism         & 0.0275      & 0.0310  \\
transportation  & 0.0547      & 0.0000  \\
\bottomrule
\end{tabular}
\label{tab:NYC_pickup_feature_correlations}
\end{table}

\begin{table}[h!]
\centering
\caption{Correlation and p-value of NYC drop-off features.}
\begin{tabular}{ccc}
\toprule
Features         & Correlation & p-value \\
\midrule
hour            & 0.5700      & 0.0000  \\
residential     & 0.0623      & 0.0000  \\
commercial      & 0.0114      & 0.3720  \\
recreation      & -0.0346     & 0.0066  \\
tourism         & -0.0359     & 0.0049  \\
transportation  & 0.0082      & 0.5190  \\
\bottomrule
\end{tabular}
\label{tab:NYC_dropoff_feature_correlations}
\end{table}

\begin{table}[h!]
\centering
\caption{Correlation and p-value of Chicago pick-up features.}
\begin{tabular}{ccc}
\toprule
Features         & Correlation & p-value \\
\midrule
hour            & 0.2225      & 0.0000  \\
residential     & 0.0081      & 0.6955  \\
commercial      & -0.0053     & 0.7996  \\
recreation      & -0.0032     & 0.8786  \\
tourism         & 0.0198      & 0.3380  \\
transportation  & -0.0026     & 0.9015  \\
\bottomrule
\end{tabular}
\label{tab:Chicago_pickup_feature_correlations}
\end{table}

\begin{table}[h!]
\centering
\caption{Correlation and p-value of Chicago drop-off features.}
\begin{tabular}{ccc}
\toprule
Features         & Correlation & p-value \\
\midrule
hour            & 0.4371      & 0.0000  \\
residential     & 0.0180      & 0.3845  \\
commercial      & -0.0518     & 0.0124  \\
recreation      & -0.0311     & 0.1332  \\
tourism         & -0.0302     & 0.1451  \\
transportation  & -0.0507     & 0.0142  \\
\bottomrule
\end{tabular}
\label{tab:Chicago_dropoff_feature_correlations}
\end{table}

\begin{table}[h!]
\centering
\caption{Correlation and p-value of DC pick-up features.}
\begin{tabular}{ccc}
\toprule
Features         & Correlation & p-value \\
\midrule
hour            & 0.2870      & 0.0000  \\
residential     & -0.0325      & 0.0707  \\
commercial      & 0.0645     & 0.0003  \\
recreation      & 0.0269     & 0.1351  \\
tourism         & 0.0688      & 0.0001  \\
transportation  & 0.0613    & 0.0006 \\
\bottomrule
\end{tabular}
\label{tab:DC_pickup_feature_correlations}
\end{table}

\begin{table}[h!]
\centering
\caption{Correlation and p-value of DC drop-off features.}
\begin{tabular}{ccc}
\toprule
Features        & Correlation & p-value \\
\midrule
hour            & 0.6585      & 0.0000  \\
residential     & 0.0099     & 0.5827  \\
commercial      & -0.0218     & 0.2243  \\
recreation      & -0.0435    & 0.0156  \\
tourism         & -0.0406     & 0.0238  \\
transportation  & -0.0125     & 0.4876  \\
\bottomrule
\end{tabular}
\label{tab:DC_dropoff_eature_correlations}
\end{table}

\subsection{[35-38] Crime}

For this experiment, we utilize crime data from New York City, Chicago, Washington D.C., and Cincinnati, with all datasets normalized to a scale ranging from 0.0 to 10.0. The New York City data comes from the NYPD Arrests Data (Historic) for 2021~\cite{nypdarrestdata}. For Chicago, we use the Chicago Crimes-2014 dataset~\cite{chicagocrimes2014}. The Washington D.C. data is derived from the Washington D.C. Crime Incidents - 2020 dataset~\cite{DCcrimeincidents2020}. Finally, for Cincinnati, we rely on the PDI (Police Data Initiative) Crime Incidents in 2023 dataset~\cite{cincinnaticrimeincidents2023}.

\subsubsection{Data Preprocess}

The crime counts are aggregated by ``zone'' and ``year'' for each city to obtain the total  ``crime count'' for each zone in the specified year. The zone could be either community or zip code. The crime rate is then calculated by normalizing the crime count to a scale of 0.0 to 10.0, which is used as the target variable for evaluation.

\subsubsection{Feature}
Unlike other tasks where features are selected by the LLM, crime features in this task are derived from demographic and POI factors as identified in \cite{wang2016crime}, which analyzed crime factors in Chicago. Note that the last feature  ``Nightlife'' is only used in Chicago to compare with previous work \cite{wang2016crime}, while the other 3 cities use the first 6 features.
\begin{itemize}
    \item Total Population: What is the total population of the area? (0: very low, 10: very high)
    \item Population Density: How densely populated is the area? (0: very sparsely populated, 10: very densely populated)
    \item Poverty Index: What is the level of poverty in the area? (0: no poverty, 10: extreme poverty)
    \item Disadvantage Index: How disadvantaged is the area compared to others? (0: not disadvantaged, 10: extremely disadvantaged)
    \item Residential Stability: How stable is the residential population in the area? (0: very unstable, 10: very stable)
    \item Ethnic Diversity: How ethnically diverse is the area? (0: not diverse at all, 10: very diverse)
    \item Nightlife: How active is the nightlife in the area? (0: very quiet, 10: very active) 
\end{itemize}



\begin{table}[h!]
\centering
\caption{Correlation and p-value of NYC crime features.}
\begin{tabular}{ccc}
\toprule
Features                    & Correlation & p-value        \\ 
\midrule
total\_population         & 0.4405      & 0.0000   \\ 
population\_density       & 0.3299      & 0.0000   \\ 
poverty\_index            & 0.5564      & 0.0000   \\ 
disadvantage\_index       & 0.5523      & 0.0000   \\ 
residential\_stability    & -0.4562     & 0.0000   \\ 
ethnic\_diversity         & 0.3668      & 0.0000   \\ 

\bottomrule
\end{tabular}
\label{tab:NYC_demographic_features}
\end{table}

\begin{table}[h!]
\centering
\caption{Correlation and p-value of Chicago crime features.}
\begin{tabular}{ccc}
\toprule
Features                    & Correlation & p-value        \\ 
\midrule
total\_population         & 0.0023      & 0.9840   \\ 
population\_density       & 0.0771      & 0.5050   \\ 
poverty\_index            & 0.6395      & 0.0000   \\ 
disadvantage\_index       & 0.7095      & 0.0000   \\ 
residential\_stability   & -0.5687     & 0.0000   \\ 
ethnic\_diversity         & -0.2503     & 0.0281   \\ 
nightlife                 & -0.2891     & 0.0108   \\
\bottomrule
\end{tabular}
\label{tab:Chicago_demographic_features}
\end{table}

\begin{table}[h!]
\centering
\caption{Correlation and p-value of Washington DC crime features.}
\begin{tabular}{ccc}
\toprule
Features                    & Correlation & p-value        \\ 
\midrule
total\_population         & 0.0196      & 0.9034   \\ 
population\_density       & 0.1218      & 0.4480   \\ 
poverty\_index            & 0.0614      & 0.7030   \\ 
disadvantage\_index       & 0.0840      & 0.6016   \\ 
residential\_stability    & -0.1581     & 0.3236   \\ 
ethnic\_diversity         & 0.2294      & 0.1491   \\ 

\bottomrule
\end{tabular}
\label{tab:DC_demographic_features}
\end{table}

\begin{table}[h!]
\centering
\caption{Correlation and p-value of Cincinnati crime features.}
\begin{tabular}{ccc}
\toprule
Features                  & Correlation & p-value  \\ 
\midrule
total\_population         & 0.4192      & 0.0004   \\ 
population\_density       & 0.3295      & 0.0065   \\ 
poverty\_index            & 0.0584      & 0.6388   \\ 
disadvantage\_index       & 0.0637      & 0.6088   \\ 
residential\_stability    & -0.1091     & 0.3794   \\ 
ethnic\_diversity         & 0.3036      & 0.0125   \\ 
\bottomrule
\end{tabular}
\label{tab:Cincinnati_demographic_features}
\end{table}

\subsection{[39-41] 311} 
\subsubsection{Data Description}

For this experiment, we use 311 service request data from New York City, Austin, and Cincinnati, with all datasets normalized to a count scale ranging from 0.0 to 10.0. The New York City data is sourced from the 311 Service Requests dataset, collected in 2023 and hosted on the NYC Open Data portal~\cite{NYC311_2023}. For Austin, the experiment uses the Austin 311 Service Requests dataset, collected in 2024 and hosted on Austin's Open Data portal~\cite{Austin311_2024}. The Cincinnati data comes from the Cincinnati 311 Non-Emergency Service Requests dataset, collected in 2024 and hosted on Cincinnati's Open Data portal~\cite{Cincinnati311_2024}.

\subsubsection{Data Preprocess}
We calculate the total number of 311 service requests made in 2022 to serve as the target variable for our model's predictions. Initially, we remove samples with nonexistent or incorrect zip codes and excluded entries where the creation time did not fall within 2022. Subsequently, we compute the count of 311 service requests for each zip code area and normalized these counts to a scale of 0.0 to 10.0 to be used as the target column.

\subsubsection{Feature}
\begin{itemize}
    \item Population Density: Measures the population density of the area (0.0: very low density, 10.0: very high density). Based on the number of residents per square mile.
    \item Unemployment Rate: Reflects the unemployment rate in the area (0.0: very low unemployment, 10.0: very high unemployment). Based on the proportion of unemployed individuals in the total labor force.
    \item Annual Income: Reflects the annual income level of residents (0.0: very low income, 10.0: very high income). Based on median household income.
    \item Crime Rate: Measures the crime rate in the area (0.0: very low, 10.0: very high).
    \item Infrastructure: Measures the quality and availability of infrastructure (0.0: very poor, 10.0: excellent). Includes transportation, utilities, and public services.
\end{itemize}

\begin{table}[h]
\centering
\caption{Correlation and p-value of New York 311 features.}
\begin{tabular}{ccc}
\toprule
Features             & Correlation & p-value        \\ 
\midrule
population\_density  & 0.2887      & 0.0001         \\ 
unemployment\_rate   & 0.3369      & 0.0000         \\ 
annual\_income       & -0.4603     & 0.0000         \\ 
infrastructure       & -0.3043     & 0.0000         \\ 
crime\_rate          & 0.4141      & 0.0000         \\ 
\bottomrule
\end{tabular}
\label{tab:NYC_311}
\end{table}

\begin{table}[h]
\centering
\caption{Correlation and p-value of Austin 311 features.}
\begin{tabular}{ccc}
\toprule
Features             & Correlation & p-value        \\ 
\midrule
population\_density  & 0.5582      & 0.0000         \\ 
unemployment\_rate   & 0.3436      & 0.0038         \\ 
annual\_income       & -0.1830     & 0.1323         \\ 
infrastructure       & 0.2195      & 0.0700         \\ 
crime\_rate          & 0.6544      & 0.0000         \\ 
\bottomrule
\end{tabular}
\label{tab:Austin_311}
\end{table}

\begin{table}[h]
\centering
\caption{Correlation and p-value of Cincinnati 311 features.}
\begin{tabular}{ccc}
\toprule
Features             & Correlation & p-value        \\ 
\midrule
population\_density  & 0.4067      & 0.0046         \\ 
unemployment\_rate   & 0.4282      & 0.0027         \\ 
annual\_income       & -0.5347     & 0.0001         \\ 
infrastructure       & -0.4306     & 0.0025         \\ 
crime\_rate          & 0.7024      & 0.0000         \\ 
\bottomrule
\end{tabular}
\label{tab:Cincinnati_311}
\end{table}



\end{document}